%% file: main.tex
\documentclass[10pt,twocolumn,letterpaper]{article}

\usepackage[pagenumbers]{cvpr} 
\usepackage{multirow}
\definecolor{cvprblue}{rgb}{0.21,0.49,0.74}
\definecolor{lighterblue}{RGB}{220, 235, 250}
\usepackage{colortbl}

\usepackage[pagebackref,breaklinks,colorlinks,allcolors=cvprblue]{hyperref}

\def\paperID{64}
\def\confName{CVPR}
\def\confYear{2025}

\title{NTIRE 2025 Challenge on Cross-Domain Few-Shot Object Detection: \\ Methods and Results}

\author{
\parbox{\textwidth}{
\centering
Yuqian Fu\textsuperscript{*} \quad Xingyu Qiu\textsuperscript{*} \quad Bin Ren\textsuperscript{*} \quad Yanwei Fu\textsuperscript{*} \quad Radu Timofte\textsuperscript{*} \quad Nicu Sebe\textsuperscript{*} 
\\ Ming-Hsuan Yang\textsuperscript{*} \quad Luc Van Gool\textsuperscript{*} \quad Kaijin Zhang \quad Qingpeng Nong \quad Xiugang Dong \\ Hong Gao \quad Xiangsheng Zhou \quad
Jiancheng Pan \quad Yanxing Liu \quad Xiao He \quad Jiahao Li \\ Yuze Sun \quad Xiaomeng Huang \quad
Zhenyu Zhang \quad Ran Ma \quad Yuhan Liu \quad Zijian Zhuang \\ Shuai Yi \quad Yixiong Zou \quad
Lingyi Hong \quad Mingxi Chen \quad Runze Li \quad Xingdong Sheng \\ Wenqiang Zhang \quad
Weisen Chen \quad Yongxin Yan \quad Xinguo Chen \quad Yuanjie Shao \\ Zhengrong Zuo \quad Nong Sang \quad
Hao Wu \quad Haoran Sun \quad Shuming Hu \quad Yan Zhang \\ Zhiguang Shi \quad Yu Zhang \quad Chao Chen \quad Tao Wang \quad Da Feng \quad Linhai Zhuo \\ Ziming Lin \quad
Yali Huang \quad Jie Me \quad Yiming Yang \quad Mi Guo \quad Mingyuan Jiu \\ Mingliang Xu \quad
Maomao Xiong \quad Qunshu Zhang \quad Xinyu Cao \quad
Yuqing Yang \\
Dianmo Sheng \quad Xuanpu Zhao \quad Zhiyu Li \quad Xuyang Ding \quad
Wenqian Li
}
}

\begin{document}
\maketitle

\begingroup
\renewcommand\thefootnote{*}
\footnotetext{Yuqian Fu, Xingyu Qiu, Bin Ren, Yanwei Fu, Radu Timofte, Nicu Sebe, Ming-Hsuan Yang, and Luc Van Gool are the NTIRE2025 challenge organizers. The other authors are participants in this challenge. \\ 
Appendix~\ref{sec:teams} contains the authors’ team names and affiliations. \\
NTIRE2025 webpage: \href{https://cvlai.net/ntire/2025/}{https://cvlai.net/ntire/2025/}. \\
Challenge Codes: \href{https://github.com/lovelyqian/NTIRE2025\_CDFSOD}{https://github.com/lovelyqian/NTIRE2025\_CDFSOD}.  
} 
\endgroup

\begin{abstract}
Cross-Domain Few-Shot Object Detection (CD-FSOD) poses significant challenges to existing object detection and few-shot detection models when applied across domains. In conjunction with NTIRE 2025, we organized the 1st CD-FSOD Challenge, aiming to advance the performance of current object detectors on entirely novel target domains with only limited labeled data. The challenge attracted 152 registered participants, received submissions from 42 teams, and concluded with 13 teams making valid final submissions.
Participants approached the task from diverse perspectives, proposing novel models that achieved new state-of-the-art (SOTA) results under both open-source and closed-source settings. In this report, we present an overview of the 1st NTIRE 2025 CD-FSOD Challenge, highlighting the proposed solutions and summarizing the results submitted by the participants.
\end{abstract}  

\section{Introduction}
\label{sec:introduction}
Few-shot object detection (FSOD)~\cite{kohler2023few} aims at allowing models to detect novel objects using minimal labeled examples. While significant progress has been made, existing FSOD methods~\cite{shangguan2023identification, qiao2021defrcn, sun2021fsce, wang2020frustratingly, shangguan2024improved, zhang2023detect} typically assume that the training (source) and testing (target) data are drawn from the same domain. However, this assumption rarely holds in real-world applications. For instance, a model trained on natural images such as those in MS-COCO~\cite{lin2014microsoft} may face substantial challenges when applied to a novel domain like remote sensing imagery. This cross-domain few-shot learning (CD-FSL) problem has attracted considerable attention in the context of classification~\cite{tseng2020cross, guo2020broader, fu2021meta, fu2022me,  zhang2022free, li2022cross, fu2023styleadv, ren2023masked, zhuo2022tgdm, tang2022learning, zha2023boosting, ren2024sharing, zhuo2024unified}, whereas its extension to object detection—i.e., cross-domain few-shot object detection (CD-FSOD)—remains much less explored.

Upon gaping at this gap, one recent work, CD-ViTO~\cite{fu2024cross}, reveals that the different object detection datasets exhibit various characters in style, inter-class variance (ICV), and indefinable boundaries (IB).  To further investigate how these factors affect the CD-FSOD, CD-ViTO thus proposes a new benchmark which takes MS-COCO as the source domain and six distinct datasets with diverse style, ICV, IB as unseen targets. Results indicate that the prior detectors all fail to generalize to those targets when the domain gap issue is observed.

To further promote the advances on CD-FSOD, we newly introduce three more unseen targets, DeepFruits~\cite{sa2016deepfruits}, Carpk~\cite{hsieh2017drone}, and CarDD~\cite{wang2023cardd} as testbeds for the CD-FSOD detectors. Following the observations in CD-ViTO, these three targets have domains different from the source data, with varying styles, ICV, and IB. Furthermore, to maximally boost the performance of models, we define the task setting proposed in CD-ViTO as \textit{closed-source CD-FSOD}, while further introducing the new  \textit{open-source CD-FSOD} setting. To be specific, the closed-source setting means the source data for model training is strictly limited, e.g., MS-COCO as in CD-ViTO; while the open-source setting relaxes this limitation and allows the participants to leverage diverse knowledge sources and foundation models to explore the upper bound on the target domains. 

In collaboration with the 2025 New Trends in Image Restoration and Enhancement (NTIRE 2025) Workshop, which is particularly interested in the model robustness under changing conditions, we present the 1st CD-FSOD Challenge. It features an open-source CD-FSOD as the main track and a closed-source CD-FSOD as a special track. For the closed-source track, MS-COCO serves as the sole source domain. The validation phase includes six target domains proposed in CD-ViTO. Three additional novel domains are used as the final test sets for both tracks. Mean Average Precision (mAP) is employed as the ranking metric. We believe this challenge will drive progress in the CD-FSOD field and foster meaningful algorithmic innovations.



This challenge is one of the NTIRE 2025\footnote{\url{https://www.cvlai.net/ntire/2025/}} Workshop associated challenges on: ambient lighting normalization~\cite{ntire2025ambient}, reflection removal in the wild~\cite{ntire2025reflection}, shadow removal~\cite{ntire2025shadow}, event-based image deblurring~\cite{ntire2025event}, image denoising~\cite{ntire2025denoising}, XGC quality assessment~\cite{ntire2025xgc}, UGC video enhancement~\cite{ntire2025ugc}, night photography rendering~\cite{ntire2025night}, image super-resolution (x4)~\cite{ntire2025srx4}, real-world face restoration~\cite{ntire2025face}, efficient super-resolution~\cite{ntire2025esr}, HR depth estimation~\cite{ntire2025hrdepth}, efficient burst HDR and restoration~\cite{ntire2025ebhdr}, cross-domain few-shot object detection~\cite{ntire2025cross}, short-form UGC video quality assessment and enhancement~\cite{ntire2025shortugc,ntire2025shortugc_data}, text to image generation model quality assessment~\cite{ntire2025text}, day and night raindrop removal for dual-focused images~\cite{ntire2025day}, video quality assessment for video conferencing~\cite{ntire2025vqe}, low light image enhancement~\cite{ntire2025lowlight}, light field super-resolution~\cite{ntire2025lightfield}, restore any image model (RAIM) in the wild~\cite{ntire2025raim}, raw restoration and super-resolution~\cite{ntire2025raw}, and raw reconstruction from RGB on smartphones~\cite{ntire2025rawrgb}.

\section{NTIRE 2025 CD-FSOD Challenge}
\subsection{Challenge Overview} \label{sec:overview}
\begin{figure*}[h]
\centering	{\includegraphics[width=1.\linewidth]{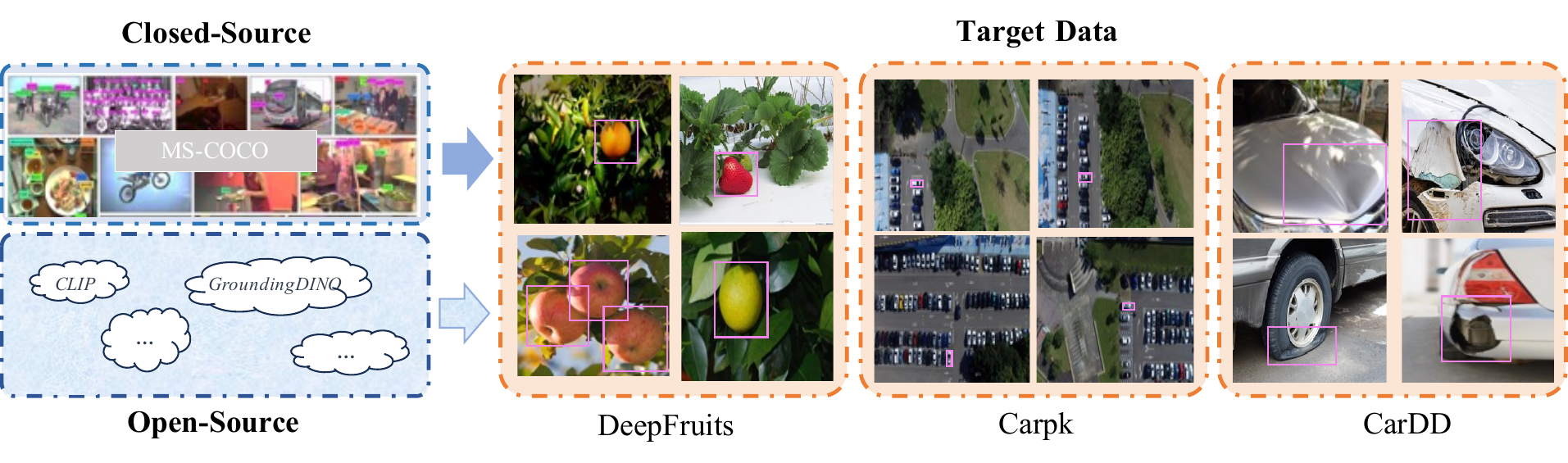}}
\caption{Illustration of the challenge settings, including the closed-source and open-source CD-FSOD tracks. The three newly introduced target datasets used in the final testing phase are also shown. \label{fig:task}}
\end{figure*}

Our challenge aims to advance \textbf{Cross-Domain Few-Shot Object Detection (CD-FSOD)} — detecting objects under domain shifts with limited labeled data. We use six previously published target domains~\cite{fu2024cross} as validation sets and introduce three newly constructed datasets for final testing. Beyond the dataset update, we introduce \textit{open-source CD-FSOD} as a new setting, allowing participants to freely choose source datasets and pre-trained models to enhance generalization. Fig.~\ref{fig:task} illustrates both the predefined closed-source CD-FSOD and the new open-source CD-FSOD settings, along with the newly introduced target domains.

\subsection{Task Formulations}
\noindent\textbf{Closed-Source CD-FSOD.} Given a source dataset $\mathcal{D}_{S}$ and a novel target dataset $\mathcal{D}_{T}$, the closed-source CD-FSOD track assumes that the source class set $\mathcal{C}_{S}$ and the target class set $\mathcal{C}_{T}$ are completely disjoint, i.e., $\mathcal{C}_{S} \cap \mathcal{C}_{T} = \emptyset$. Additionally, the distributions of the source domain $\mathcal{D}_{S}$ and the target domain $\mathcal{D}_{T}$ are not identical. Participants are required to train models on $\mathcal{D}_{S}$ and test them on $\mathcal{D}_{T}$, where each class in $\mathcal{C}_{T}$ has only a few labeled examples. Usually, $\mathcal{D}_{S}$ is a single dataset, as in CD-ViTO~\cite{fu2024cross}. We refer to this setting as closed-source CD-FSOD to differentiate it from the open-source variant.

\noindent\textbf{Open-Source CD-FSOD.} In contrast to the closed-source setting where training data is strictly limited, the open-source CD-FSOD track is designed to leverage the capabilities of foundation models. Since these models are pretrained on large-scale and diverse datasets, it is practically hard to trace all the knowledge embedded within them.
Hence, we refer to this setting as \textit{open-source}. While the relaxed constraints on source data make it difficult to strictly ensure non-overlapping classes between the source and target data, the track still focuses on addressing the core challenges of domain shift and few-shot object detection. We believe this setting will significantly accelerate the development of CD-FSOD methods for real-world applications.

In this challenge, the open-source CD-FSOD is designated as the main track, with awards presented to the top three teams. The closed-source CD-FSOD serves as the special track, with a single award granted to the top-performing team.

\noindent \textbf{$N$-way $K$-shot Protocol.} We adopt the $N$-way $K$-shot evaluation protocol. For each novel class in the target class set $\mathcal{C}_{T}$, $K$ labeled instances are provided, forming the support set $S$. The remaining unlabeled instances constitute the query set $Q$. Instances contained in the support set $S$ are used to assist the model in recognizing and detecting the objects in $Q$.

\subsection{Challenge Phases and Datasets}
This challenge involves one development stage and one testing stage.  The source data $\mathcal{D}_{S}$ for both stages is the same, i.e., MS-COCO~\cite{lin2014microsoft} for the closed-source track and unlimited data for the open-source track. While the testing data $\mathcal{D}_{T}$ is different. 

\noindent\textbf{Development Stage:} Datasets proposed in the CD-ViTO, including ArTaxOr~\cite{GeirArTaxOr}, Clipart1K~\cite{inoue2018cross}, DIOR~\cite{li2020object}, DeepFish~\cite{saleh2020realistic}, NEU-DET~\cite{song2013noise}, and UODD~\cite{jiang2021underwater} are taken as targets $\mathcal{D}_{T}$ during development stage. 

\noindent\textbf{Testing Stage.} Three previously unseen datasets (DeepFruits~\cite{sa2016deepfruits}, Carpk~\cite{hsieh2017drone}, and CarDD~\cite{wang2023cardd}) are introduced and used as the targets $\mathcal{D}_{T}$ for the final testing phase. Note that the ground truth annotations for these query sets are held exclusively by the challenge organizers.

\subsection{CD-ViTO Baseline Model}
\label{sec:baseline_model}
\begin{figure}[h]
\centering	{\includegraphics[width=1.\linewidth]{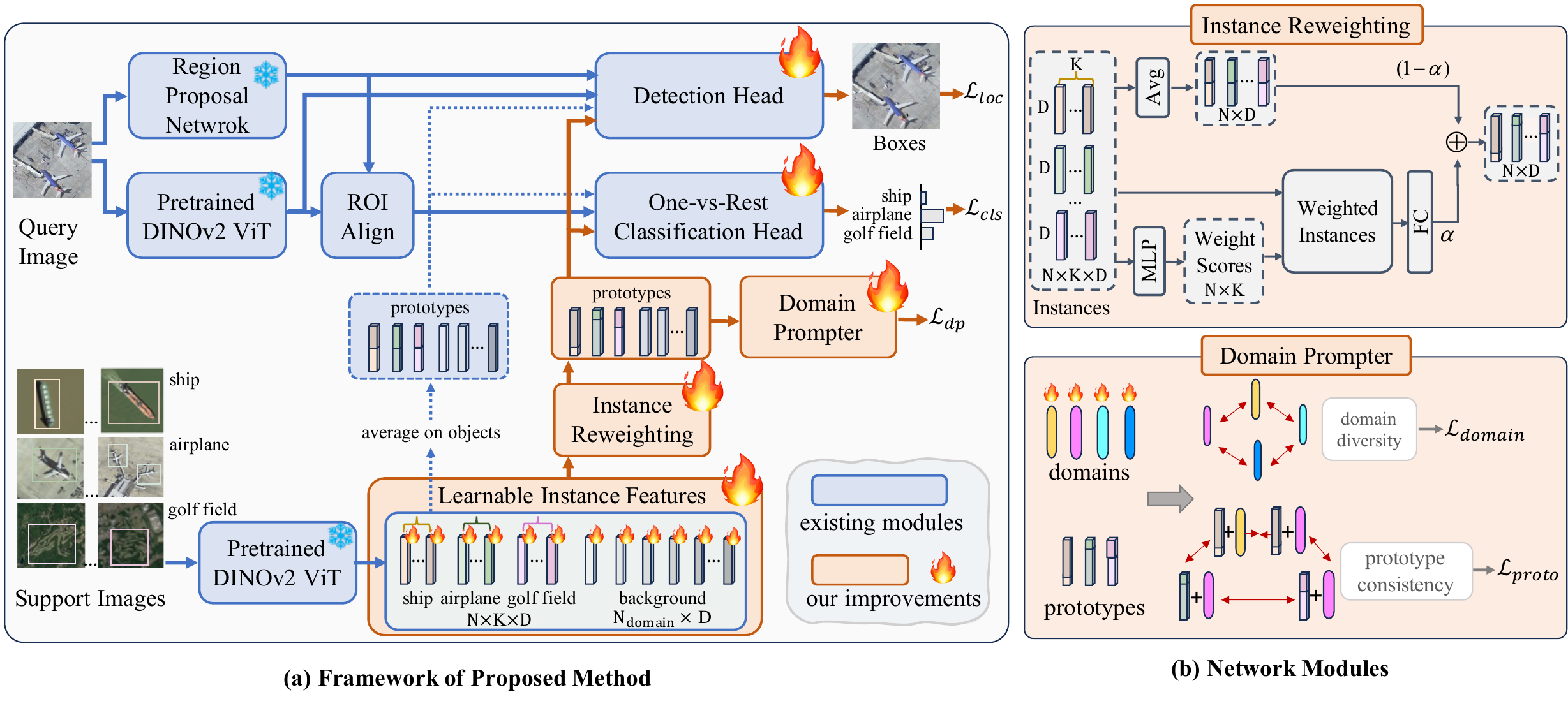}}
\vspace{-0.15in}
\caption{Overall framework of CD-ViTO baseline method. \label{fig:framework-base} 
}
\end{figure}

We take CD-ViTO, the current State-of-the-art (SOTA) method under the closed-source setting, as the baseline for this challenge. Briefly, CD-ViTO is built upon DE-ViT~\cite{zhang2023detect}, an open-set detector, and fine-tuned using the support set. As in Fig.~\ref{fig:framework-base}, modules in blue are inherited from DE-ViT, while modules in orange are newly proposed. New improvements include 
\textit{learnable instance features}, \textit{instance reweighting}, 
\textit{domain prompter}, and \textit{finetuning}.  

Intuitively, the learnable instance feature module is designed to enhance inter-class variance (ICV) among different target classes by making the initially fixed instance features learnable and optimizing them through supervised few-shot detection tasks on the target support set. The instance reweighting module further improves prototype quality by assigning higher weights to high-quality object instances—e.g., those with minimal indefinable boundary (IB). These weights are learned via a lightweight MLP and fully connected layer, as illustrated in the upper part of Fig.~\ref{fig:framework-base}(b). The domain prompter module introduces learnable domain perturbations to simulate varying domain styles. These perturbations are applied to object prototypes, followed by a prototype consistency loss to ensure that the introduced perturbations do not affect the semantic category of the prototypes. Simultaneously, a domain diversity loss encourages the generated domains to be sufficiently diverse. 
The lower part of Fig.~\ref{fig:framework-base}(b) illustrates this mechanism. By injecting virtual domains and enforcing robustness against the induced perturbations, this strategy enhances the model’s generalization under domain shifts.
Finetuning is applied to the modules highlighted with fire icons in Fig.~\ref{fig:framework-base}.

\subsection{Evaluation Protocol}
The final score is measured based on the model's performance on the three datasets of the testing stage. For each dataset, we validate the models on three different few-shot settings: 1-shot, 5-shot, and 10-shot. This results in a total of nine mean Average Precision (mAP) scores: \texttt{D1\_1shot}, \texttt{D1\_5shot}, \texttt{D1\_10shot}; \texttt{D2\_1shot}, \texttt{D2\_5shot}, \texttt{D2\_10shot}; and \texttt{D3\_1shot}, \texttt{D3\_5shot}, \texttt{D3\_10shot}. The \texttt{D1, D2, D3} denote the Deep-Fruits, Carpk, and CarDD, respectively. 

The final ranking score is computed as a weighted average $avg()$ of these scores:
\begin{align*}\small
\textit{Score} =\ & 2*\textit{avg}(\texttt{D1\_1shot}, \texttt{D2\_1shot}, \texttt{D3\_1shot}) \\
&+ 1*\textit{avg}(\texttt{D1\_5shot}, \texttt{D2\_5shot}, \texttt{D3\_5shot}) \\
&+ 1*\textit{avg}(\texttt{D1\_10shot}, \texttt{D2\_10shot}, \texttt{D3\_10shot})
\end{align*}

\noindent\textbf{Rationale for Weighted Scoring.}
We assign a higher weight (×2) to the 1-shot setting for two primary reasons: (1) Performance in the 1-shot scenario is generally lower than in the 5-shot and 10-shot settings due to the limited availability of labeled examples for adaptation; and (2) emphasizing 1-shot performance encourages the development of models that are more robust and effective in extremely low-data conditions.

\begin{table*}[ht]
\centering
\caption{Open-source and closed-source results on CD-FSOD. D1, D2, and D3 represent DeepFruits, CARPK, and CarDD, respectively. Mean Average Precision (mAP) on 1-shot, 5-shot, and 10-shot are reported. Teams achieving top results are highlighted.} \label{tab:result}
\resizebox{\linewidth}{!}{  
\begin{tabular}{|c|l|c|c|c|c|c|c|c|c|c|c|}
\hline
\multicolumn{12}{|c|}{ \cellcolor{cvprblue!50} \textbf{Main Open-Source Track}} \\
\hline
\textbf{Rank} & \textbf{Team Name} & \textbf{Score} & \textbf{D1\_1shot} & \textbf{D1\_5shot} & \textbf{D1\_10shot} & \textbf{D2\_1shot} & \textbf{D2\_5shot} & \textbf{D2\_10shot} & \textbf{D3\_1shot} & \textbf{D3\_5shot} & \textbf{D3\_10shot} \\
\hline
\cellcolor{cvprblue!15}1 & \cellcolor{cvprblue!15}MoveFree& \cellcolor{cvprblue!15}\textbf{231.01} & \cellcolor{cvprblue!15}\textbf{66.18} & \cellcolor{cvprblue!15}64.58 & \cellcolor{cvprblue!15}62.57 & \cellcolor{cvprblue!15}60.43 & \cellcolor{cvprblue!15}58.89 & \cellcolor{cvprblue!15}59.00 & \cellcolor{cvprblue!15}\textbf{48.75} & \cellcolor{cvprblue!15}\textbf{49.28} & \cellcolor{cvprblue!15}\textbf{48.00} \\
\cellcolor{cvprblue!15}2 & \cellcolor{cvprblue!15}AI4EarthLab & \cellcolor{cvprblue!15}215.92 & \cellcolor{cvprblue!15}61.19 & \cellcolor{cvprblue!15}\textbf{65.41} & \cellcolor{cvprblue!15}\textbf{65.35} & \cellcolor{cvprblue!15}59.15 & \cellcolor{cvprblue!15}58.05 & \cellcolor{cvprblue!15}59.00 & \cellcolor{cvprblue!15}34.21 & \cellcolor{cvprblue!15}43.85 & \cellcolor{cvprblue!15}47.00 \\
\cellcolor{cvprblue!15}3 & \cellcolor{cvprblue!15}IDCFS & \cellcolor{cvprblue!15}215.48 & \cellcolor{cvprblue!15}63.34 & \cellcolor{cvprblue!15}\textbf{65.41} & \cellcolor{cvprblue!15}64.75 & \cellcolor{cvprblue!15}\textbf{61.14} & \cellcolor{cvprblue!15}60.42 & \cellcolor{cvprblue!15}\textbf{60.00} & \cellcolor{cvprblue!15}32.33 & \cellcolor{cvprblue!15}39.24 & \cellcolor{cvprblue!15}43.00 \\
4 & FDUROILab\_Lenovo & 211.55 & 61.25 & 62.89 & 64.66 & 59.24 & 59.24 & 59.00 & 35.13 & 37.63 & 40.00 \\
5 & HUSTLab & 210.78 & 63.71 & 61.32 & 57.19 & 60.42 & \textbf{60.47} & \textbf{60.00} & 31.01 & 40.09 & 43.00 \\
6 & TongjiLab & 172.14 & 42.36 & 41.90 & 41.74 & 55.95 & 55.95 & 55.00 & 31.40 & 31.40 & 31.00 \\
7 & Manifold & 159.86 & 32.05 & 44.28 & 44.27 & 57.06 & 57.06 & 57.00 & 18.71 & 29.34 & 32.00 \\
8 & MXT & 108.20 & 22.26 & 40.57 & 41.34 & 21.12 & 26.34 & 30.23 & 23.81 & 28.00 & 29.00 \\
\hline
\multicolumn{12}{|c|}{ \cellcolor{cvprblue!50}  \textbf{Special Closed-Source Track}} \\
\hline
\textbf{Rank} & \textbf{Team Name} & \textbf{Score} & \textbf{D1\_1shot} & \textbf{D1\_5shot} & \textbf{D1\_10shot} & \textbf{D2\_1shot} & \textbf{D2\_5shot} & \textbf{D2\_10shot} & \textbf{D3\_1shot} & \textbf{D3\_5shot} & \textbf{D3\_10shot} \\
\hline
\cellcolor{cvprblue!15}1 & \cellcolor{cvprblue!15}X-Few & \cellcolor{cvprblue!15}\textbf{125.90} & \cellcolor{cvprblue!15}\textbf{36.58} & \cellcolor{cvprblue!15}46.95 & \cellcolor{cvprblue!15}\textbf{50.98} & \cellcolor{cvprblue!15}\textbf{23.01} & \cellcolor{cvprblue!15}\textbf{29.68} & \cellcolor{cvprblue!15}\textbf{28.00} & \cellcolor{cvprblue!15}\textbf{20.11} & \cellcolor{cvprblue!15}29.68 & \cellcolor{cvprblue!15}33.00 \\
2 & MM & 117.39 & 32.47 & 45.23 & 50.23 & 18.83 & 29.36 & \textbf{28.00} & 18.31 & 29.14 & 31.00 \\
3 & FSV & 112.81 & 31.23 & 43.89 & 49.32 & 13.69 & 26.04 & 26.59 & 19.71 & 30.16 & \textbf{33.17} \\
4 & IPC & 105.62 & 32.58 & \textbf{47.12} & 45.64 & 13.41 & 20.77 & 13.00 & 18.18 & 29.98 & 32.00 \\
5 & LJY & 105.28 & 33.52 & 46.04 & 45.34 & 10.68 & 11.45 & 25.00 & 18.34 & \textbf{30.94} & 32.00 \\
/ & CD-ViTO Base~\cite{fu2024cross} & 91.00 & 27.95 & 37.42 & 43.58 & 6.77 & 21.28 & 24.00 & 10.07 & 26.47 & 30.00\\
\hline
\end{tabular}}
\end{table*}

\section{Challenge Results}
Among the 152 registered participants, 8 and 5 teams have participated the final testing stage and submitted their results, codes, and factsheets. Table.~\ref{tab:result} summarizes the results of these methods.  Detailed descriptions of the participants' solutions are provided in Sec.\ref{sec:teams-solution} and Sec.\ref{sec:teams-solution2}, each corresponding to a different track.

\noindent\textbf{Open-Source Track Results.} In the open-source track, nearly all participating teams achieved strong performance with clear improvements over the provided CD-ViTO baseline. This highlights not only the effectiveness of their proposed methods but also the significance of introducing this new task setting. As observed, relaxing the strict limitation on the source data offers a substantial advantage in tackling the CD-FSOD task.

Specifically, the teams MoveFree, AI4EarthLab, and IDCFS emerged as the top performers in this track, achieving scores of 231.01, 215.92, and 215.48, respectively—significantly surpassing the baseline and other teams under the same track.

\noindent\textbf{Closed-Source Track Results.} The performance achieved by the closed-source track teams is generally lower than that of the open-source track. This is quite understandable considering that the closed-source track enforces stricter constraints. Nevertheless, the participants managed to improve the baseline method clearly.

In particular, the X-Few team stands out with a final score of 125.90, significantly outperforming other competitors. This shows that well-designed architectures and training strategies can still bring notable gains even without relying on large external models. Other teams in this track also delivered solid improvements. Their contributions are valuable in terms of enabling fair comparisons and emphasizing algorithmic annotations.

\section{Main Open-Source Track Methods}
\label{sec:teams-solution}
\input{teams/team10_MoveFree/main}

\input{teams/team06_AI4EarthLab/main}

\input{teams/team15_IDCFS/main}

\input{teams/team13_FDUROILab_Lenovo/main}

\input{teams/team14_HUSTLab/main}

\input{teams/team03_TongjiLab/main}

\input{teams/team05_Manifold/main}

\input{teams/team07_MXT/main}

\section{Special Closed-Source Track Methods}
\label{sec:teams-solution2}

\input{teams/team12_X-Few/main}

\input{teams/team08_MM/main}

\input{teams/team16_FSV/main}

\input{teams/team01_IPC/main}

\input{teams/team11_LJY/main}

\section*{Acknowledgments}
INSAIT, Sofia University "St. Kliment Ohridski". Partially funded by the Ministry of Education and Science of Bulgaria’s support for INSAIT as part of the Bulgarian National Roadmap for Research Infrastructure.
This work was partially supported by the Humboldt Foundation. We thank the NTIRE 2025 sponsors: ByteDance, Meituan, Kuaishou, and University of Wurzburg (Computer Vision Lab).

\clearpage
\appendix

\section{Teams and affiliations}
\label{sec:teams}
\subsection*{NTIRE 2025 team}
\noindent\textit{\textbf{Title: }} NTIRE 2025 Challenge on Cross-Domain Few-Shot Object Detection: Methods and Results.\\
\noindent\textit{\textbf{Members: }} \\
Yuqian Fu$^{1}$ (\href{mailto:yuqian.fu@insait.ai}{yuqian.fu@insait.ai}),\\
Xingyu Qiu$^2$ (\href{mailto:xyqiu24@m.fudan.edu.cn}{xyqiu24@m.fudan.edu.cn}),\\
Bin Ren$^{3,4}$ (\href{mailto: bin.ren@unitn.it}{bin.ren@unitn.it}),\\
Yanwei Fu$^{2}$ (\href{mailto:yanweifu@fudan.edu.cn}{yanweifu@fudan.edu.cn}),\\
Radu Timofte$^{5}$ (\href{mailto:radu.timofte@uni-wuerzburg.de}{radu.timofte@uni-wuerzburg.de}),\\
Nicu Sebe$^{4}$ (\href{mailto:niculae.sebe@unitn.it}{niculae.sebe@unitn.it}),\\
Ming-Hsuan Yang$^{6}$ (\href{mailto:mhyang@ucmerced.edu}{mhyang@ucmerced.edu}),\\
Luc Van Gool$^{1}$ (\href{mailto:luc.vangool@insait.ai}{luc.vangool@insait.ai})\\
\noindent\textit{\textbf{Affiliations: }}\\
$^1$ INSAIT, Sofia University St. Kliment Ohridski, Bulgaria\\
$^2$ Fudan University, China\\
$^3$ University of Pisa, Italy\\
$^4$ University of Trento, Italy\\
$^5$ Computer Vision Lab, University of W\"urzburg, Germany\\
$^6$ University of California at Merced, United States\\

\input{teams/team10_MoveFree/affiliation}

\input{teams/team06_AI4EarthLab/affiliation}

\input{teams/team15_IDCFS/affiliation}

\input{teams/team13_FDUROILab_Lenovo/affiliation}

\input{teams/team14_HUSTLab/affiliation}

\input{teams/team03_TongjiLab/affiliation}

\input{teams/team05_Manifold/affiliation}

\input{teams/team07_MXT/affiliation}

\input{teams/team12_X-Few/affiliation}

\input{teams/team08_MM/affiliation}

\input{teams/team16_FSV/affiliation}

\input{teams/team01_IPC/affiliation}

\input{teams/team11_LJY/affiliation}

\clearpage
{
    \small
    \bibliographystyle{ieeenat_fullname}
    \bibliography{main}
}

\end{document}

%% file: teams/team10_MoveFree/main.tex
\subsection{MoveFree}
\subsubsection{Proposed Method}

Open-set object detectors, such as \cite{li2022grounded}, \cite{liu2024grounding}, and \cite{ren2024grounding}, are designed to detect objects based on arbitrary text descriptions. These models are typically pre-trained on large-scale, well-annotated datasets, ensuring strong alignment between textual and visual modalities. As a result, they exhibit remarkable zero-shot capabilities, allowing them to recognize and localize unseen object categories based solely on textual prompts. 
Given the strong generalization ability of such open-set detectors, this team believes that they are inherently well-suited for cross-domain few-shot object detection, as their robust pre-trained representations can be effectively adapted to new domains with minimal supervision.

Thus, the MoveFree team focuses on leveraging and enhancing pre-trained open-set object detectors for CD-FSOD during the fine-tuning stage. The proposed approach introduces three key improvements: (1) To address the issue of missing annotations, self-training is introduced to iteratively refine the training data, thereby enhancing fine-tuning performance. (2) A Mixture-of-Experts (MoE) architecture is integrated into the open-set object detector to improve adaptability and robustness in the few-shot setting. (3) A two-stage fine-tuning pipeline is designed carefully. Code is made available \footnote{\url{https://github.com/KAIJINZ228/Few_Shot_GD}}.

\noindent \textbf{Self-training Paradigm.} According to the definition of few-shot object detection in CD-ViTO\cite{fu2024cross}, $K$-shot object detection refers to having $K$ labeled instances in the training data, rather than $K$ fully annotated images. This implies that instances of target categories may lack annotations in the provided training set.

Upon careful investigation, this team identified that the issue of incomplete annotations is prominent across all three test datasets in this challenge. Drawing on their expertise in developing open-set object detectors, the team recognized that missing annotations for target categories can significantly degrade model performance. This degradation occurs because the loss function penalizes the model for correctly detecting unannotated objects, mistakenly treating them as false positives due to their absence in the ground truth labels.
Therefore, this team employs a self-training strategy during the fine-tuning stage of Grounding DINO to iteratively refine the annotations in the training data. Specifically, Grounding DINO periodically generates predictions on the training set, which are then incorporated as additional annotations. This iterative process gradually improves the quality of the training data, ultimately leading to enhanced model performance.

\noindent\textbf{The substitution of the Mixture-of-Experts (MoE).} In few-shot object detection, the availability of training data is highly limited. Therefore, maximizing the object detector’s ability to extract supervision from this scarce data is crucial during the fine-tuning stage. In this challenge, beyond the few-shot constraint, the cross-domain setting further increases the difficulty, as detectors usually require additional supervision to effectively adapt to a new domain.

The core concept of the MoE architecture is to enable different components (i.e., experts) of a model to specialize in different aspects of the data \cite{cai2024survey}. In recent years, MoE has gained popularity in multi-modal models, including Mistral \cite{jiang2024mixtral} and DeepSeek-V2 \cite{liu2024deepseek}. A common application of MoE in such models is replacing the traditional feed-forward network (FFN) with an MoE-based variant, as seen in Switch Transformer \cite{fedus2022switch} and OpenMoe \cite{xue2024openmoe}.

To maximize supervision and enable the model to learn effectively from the limited training data, this team integrates a Mixture-of-Experts (MoE) mechanism into Grounding DINO during the fine-tuning stage. The MoE framework allows different experts to specialize in distinct aspects of the data, facilitating the capture of more diverse and informative representations. It is hypothesized that this capability helps Grounding DINO better adapt to the target domain while making more efficient use of the available training data.

\begin{figure}[htbp]
    \centering
    \includegraphics[width=0.5\textwidth]{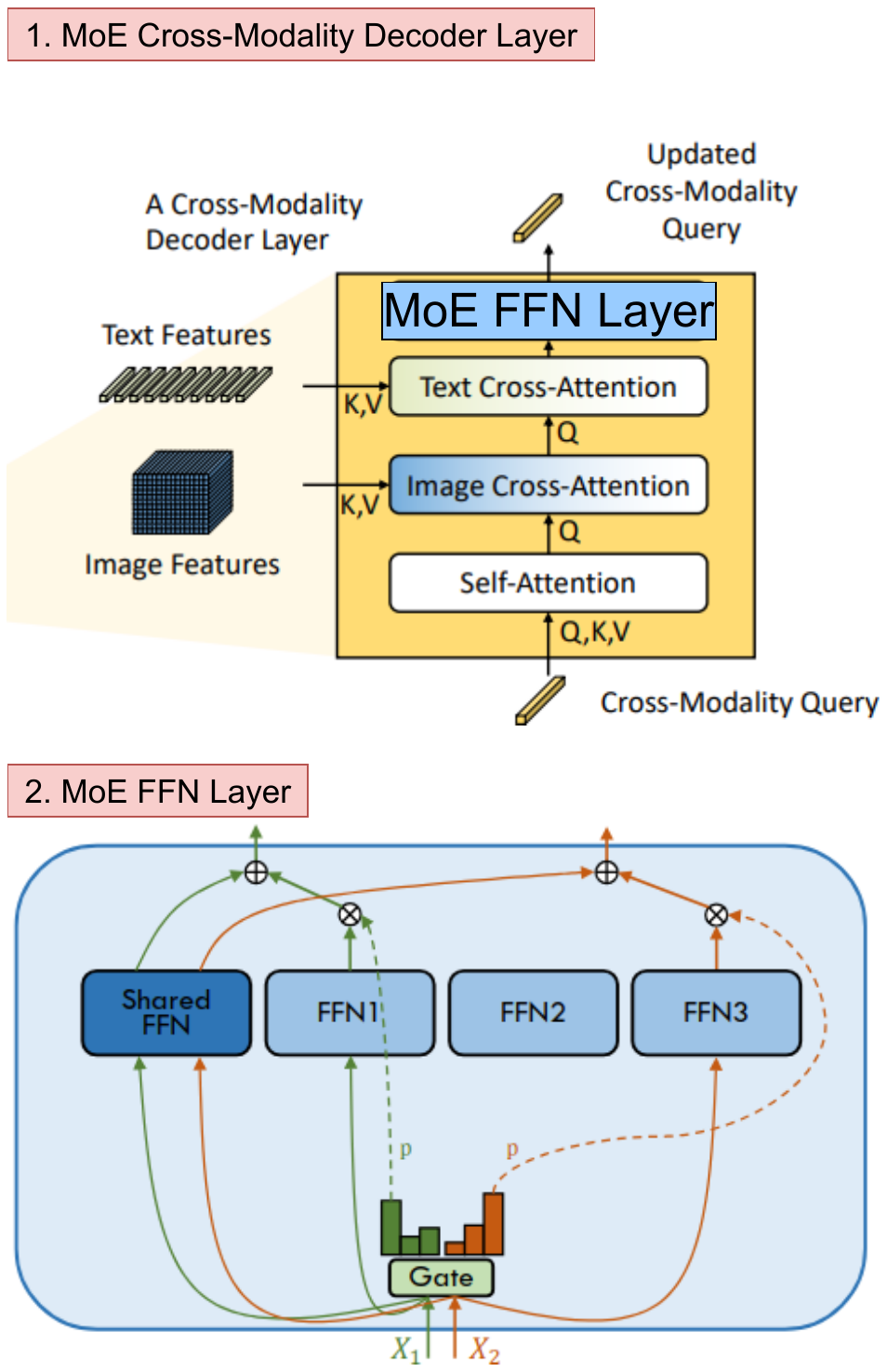}
    \vspace{-0.35in}
    \caption{Team MoveFree: an illustration of the substitution of MoE into Grounding DINO's decoder layers.}
    \label{fig:groundingdino_moe}
\end{figure}

In this team's approach, the MoE mechanism is incorporated into the feed-forward network (FFN) layers of Grounding DINO’s Cross-Modality Decoder. As illustrated in Figure~\ref{fig:groundingdino_moe}, the MoE architecture consists of one shared expert and three router-selected experts.

\subsubsection{Training Details}
A two-stage fine-tuning pipeline is adopted to adapt Grounding DINO for cross-domain few-shot object detection. In the first stage, the standard Grounding DINO (without the MoE substitution) is fine-tuned on the training data, with all parameters trainable except for the language encoder. In the second stage, the MoE architecture is introduced into the model.

For the second stage, the model is initialized using the weights obtained from the first stage, excluding the MoE components. The shared expert within the MoE is initialized with weights from the first stage, while the three router-selected experts are initialized using the open-source pre-trained weights of Grounding DINO. This initialization strategy facilitates effective learning from limited training data while retaining knowledge acquired during the initial stage. During this phase, only the MoE components and the detection head remain trainable, with all other parts of the model kept frozen.

Additionally, the self-supervised learning paradigm is applied in both stages to iteratively refine the training data and enhance performance.
The training strictly adheres to the provided few-shot training set, without utilizing any external data. The overall approach is computationally efficient and can be executed on a single V100 GPU within a reasonable time frame.

%% file: teams/team06_AI4EarthLab/main.tex
\subsection{AI4EarthLab}
\subsubsection{Proposed Method}
Foundation models pretrained on large-scale datasets, such as GroundingDINO~\cite{liu2024grounding} and LAE-DINO~\cite{pan2024locateearthadvancingopenvocabulary}, have demonstrated strong detection performance in cross-domain zero-shot and few-shot object detection tasks. Thus, the AI4EarthLab team is motivated to explore such foundation models for CD-FSOD. 

\begin{figure}[h]
\centering	{\includegraphics[width=1.\linewidth]{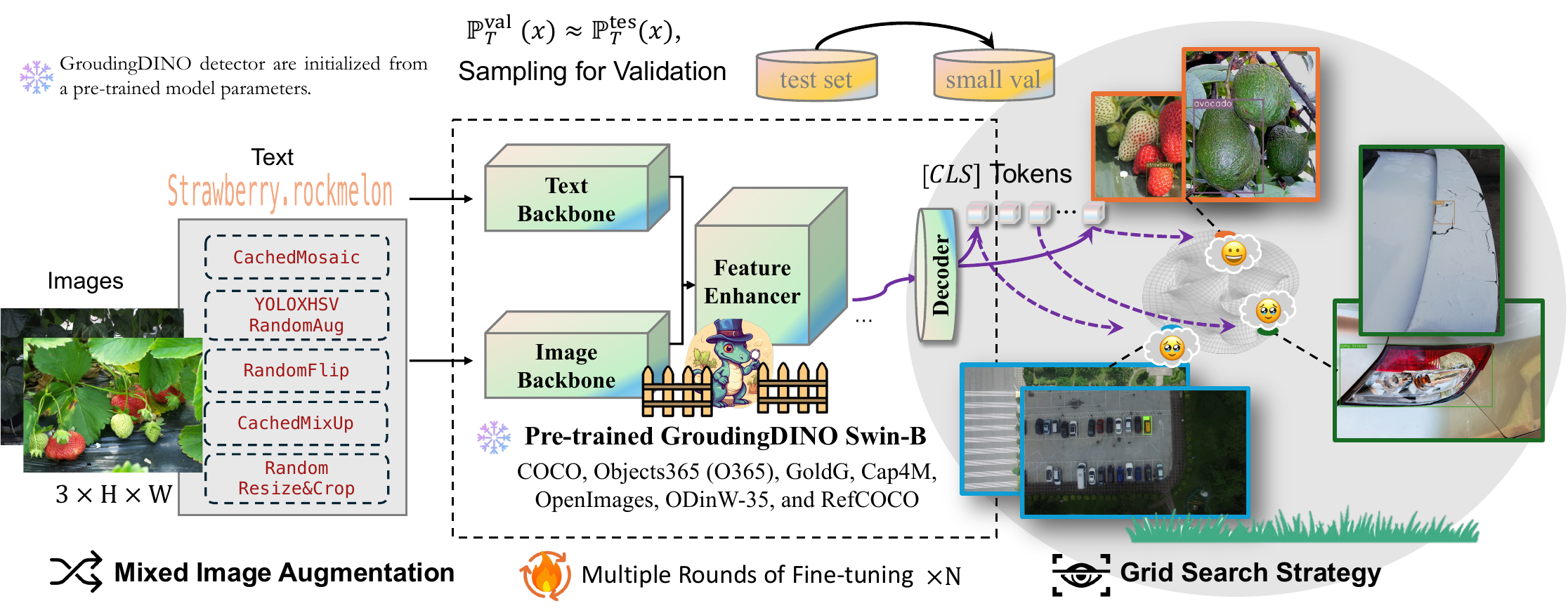}}
\caption{Team AI4EarthLab: overall framework of augmentation-search strategy Enhance Then Search (ETS) with foundation model for CD-FSOD. \label{fig:framework_ai4earth} 
}
\end{figure}

As shown in Fig.~\ref{fig:framework_ai4earth}, this team proposes an augmentation-search strategy for CD-FSOD, which leverages open-source data and transfers the model to novel target domains. Following the approaches in \cite{fu2024cross,pan2024pir}, an efficient fine-tuning method is adopted to explore the cross-domain few-shot detection capabilities of foundation models, requiring only lightweight tuning to identify effective subfields. Code is made available \footnote{\url{https://github.com/jaychempan/ETS}}.

Data augmentation has proven effective in reducing semantic confusion during few-shot fine-tuning, particularly in cases where categories—such as certain fruits—are visually and semantically similar. Through extensive few-shot experiments, it is observed that integrating image-based augmentation with optimal domain search strategies can further enhance the performance of foundation models, though their upper performance bound remains uncertain. Building upon the open-source Grounding DINO framework, several commonly used image augmentation techniques are incorporated, and specific optimization objectives are defined to efficiently search for optimal sub-domains within a broad domain space. This strategy facilitates more effective few-shot object detection.
The proposed augmentation-search strategy consists of the following steps:

\noindent \textbf{Step 1: Select the foundation model.}
This team adopts the \texttt{Swin-B} version of GroundingDINO as the foundation model, because of its best performance within the open-source model. This model has been pre-trained on a diverse set of large-scale datasets, including COCO, Objects365 (O365), GoldG, Cap4M, OpenImages, ODinW-35, and RefCOCO, which collectively provide strong generalization capabilities across multiple vision-language grounding tasks.

\noindent \textbf{Step 2: Build a combined image augmentation pipeline.} 
To improve the model's adaptability to various subdomains under limited data scenarios, this team construct a composite image augmentation pipeline. This pipeline randomly applies a combination of augmentation techniques such as \texttt{CachedMosaic}, \texttt{YOLOXHSVRandomAug}, \texttt{RandomFlip}, \texttt{CachedMixUp}, \texttt{RandomResize}, and \texttt{RandomCrop}. These methods are designed to enhance sample diversity, simulate domain shifts, and improve the model’s robustness during fine-tuning. 
Additional data augmentation techniques, such as \texttt{Copy-Paste}, are also evaluated. However, these methods are found to introduce greater instability during few-shot fine-tuning.

\noindent \textbf{Step 3: Construct an optimized target domain validation set.}
To evaluate adaptation performance, a subset of the annotated test data is sampled and used as a validation set. Rather than employing full annotations, coarse-grained labeling is applied to provide sufficient supervision for hyperparameter tuning, while significantly reducing annotation costs in the target domain.

\noindent \textbf{Step 4: Search for the best model parameters on the validation set.} 
Hyperparameter search and model selection are conducted based on validation performance. This process involves tuning the learning rate, augmentation intensity, and other training configurations to determine the optimal setup for effective domain adaptation.

\noindent \textbf{Step 5: Perform inference on the test set.} 
Once the optimal configuration is identified, the fine-tuned model is applied to the held-out test set to evaluate its final performance on the target domain.

\subsubsection{Training Details} 
Experiments are conducted on eight NVIDIA A100 GPUs, executing 8 × 50 experiment groups per round. During training, the optimal step size is selected based on historical performance to accelerate the fine-tuning process. Learning rate schedules are adjusted using milestone epochs, typically set to 1, 5, and 9 depending on the fine-tuning setting. The model uses 900 queries by default and a maximum text token length of 256. A BERT-based text encoder with BPE tokenization is employed. Both the feature enhancer and cross-modality decoder consist of six layers, and deformable attention is adopted in the image cross-attention modules. The loss function comprises classification (or contrastive) loss, box L1 loss, and GIoU loss. Following the Grounding DINO framework, Hungarian matching weights are set to 2.0 (classification), 5.0 (L1), and 2.0 (GIoU), while the final loss weights are 1.0, 5.0, and 2.0, respectively. Although various hyperparameter configurations are also explored, their impact is found to be relatively minor compared to that of data augmentation strategies.

%% file: teams/team15_IDCFS/main.tex
\subsection{IDCFS}
\subsubsection{Proposed Method}
\begin{figure}[h]
    \centering
    \includegraphics[width=1.\linewidth]{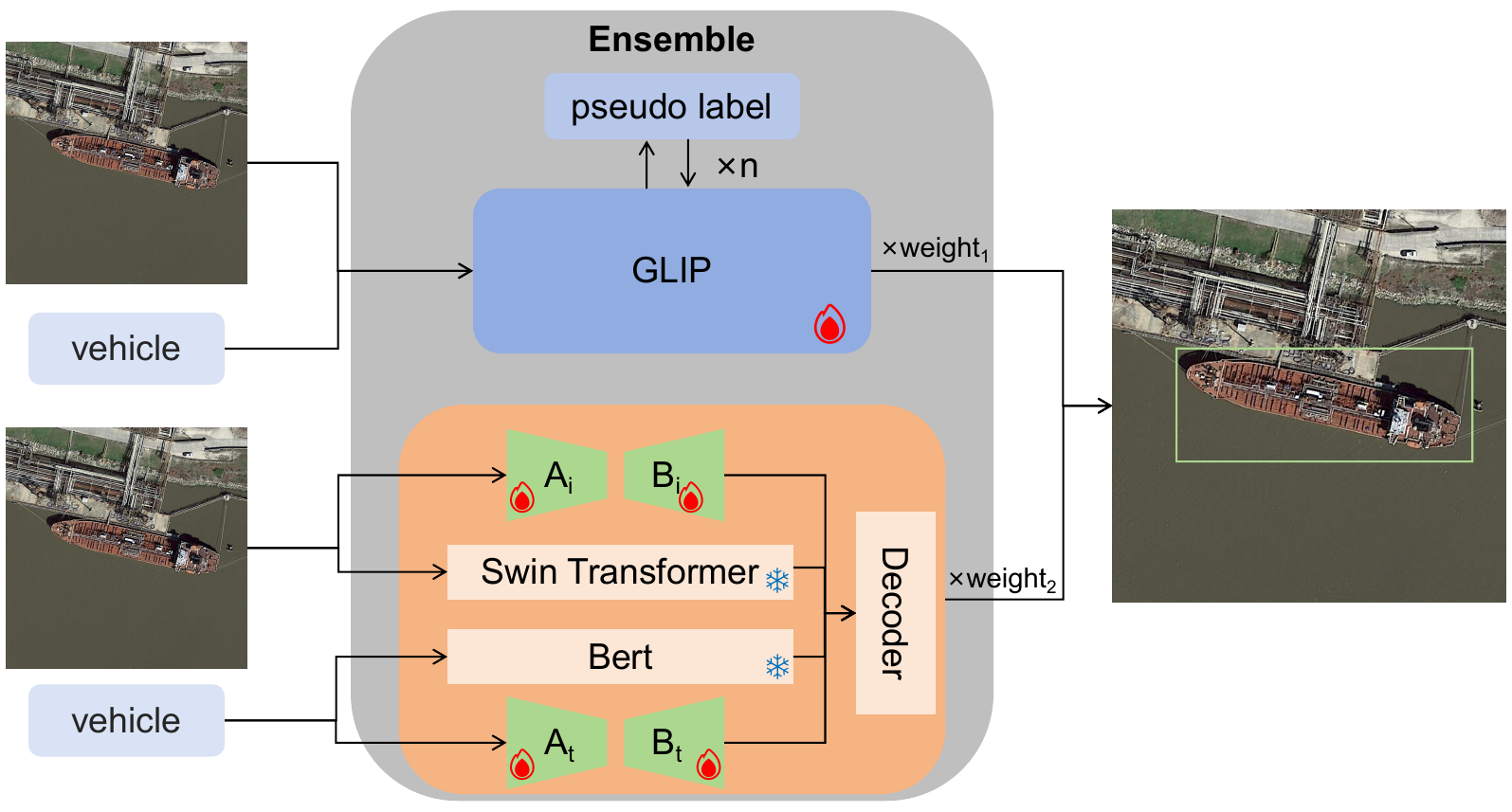} 
    \caption{Team IDCFS: overview of the proposed Pseudo-Label Driven Vision-Language Grounding for CD-FSOD.}
    \label{fig:15_method}
\end{figure}

The IDCFS team proposes a Pseudo-Label Driven Vision-Language Grounding method for CD-FSOD. As shown in Figure~\ref{fig:15_method}, the proposed method mainly combines large-scale foundation models with an iterative pseudo-labeling strategy. The GLIP~\cite{li2022grounded} is being fine-tuned using three approaches, with the full model fine-tuned delivering the best results in most cases. To better exploit the support set, an iterative training strategy is proposed and applied, using high-confidence predictions as pseudo-labels to refine the model. Additionally, this team also fine-tunes Grounding DINO~\cite{liu2024grounding} with LoRA~\cite{hu2022lora}, efficiently modifying the attention layers while freezing the base model. Finally, the model ensemble with confidence-reweighted NMS is further adopted to boost accuracy.  Code is made available \footnote{\url{https://github.com/Pumpkinder/GLIP-CDFSOD}}.

\noindent \textbf{Fine-tuning on GLIP.} Foundation models pretrained on large-scale datasets, such as GLIP~\cite{li2022grounded}, have demonstrated strong performance in zero-shot and few-shot object detection tasks. The proposed method is based on the GLIP-L model, which has been pretrained on several datasets including FourODs, GoldG, CC3M+12M, and SBU. For downstream tasks, this team tried three ways to fine-tune GLIP: 1) Full Model Fine-Tuning: fine-tune all parameters of the GLIP-L model using a relatively small learning rate (lr = 2e-5). 2) Prompt Tuning V1: fine-tune only the parameters of the text branch. 3) Prompt Tuning V2: This mode performs traditional prompt tuning by applying a linear layer to map the extracted text features.
Experiments show that Full Model Fine-Tuning generally achieves the best fine-tuning performance in most cases.

\begin{figure}[h]
    \centering
    \includegraphics[width=\linewidth]{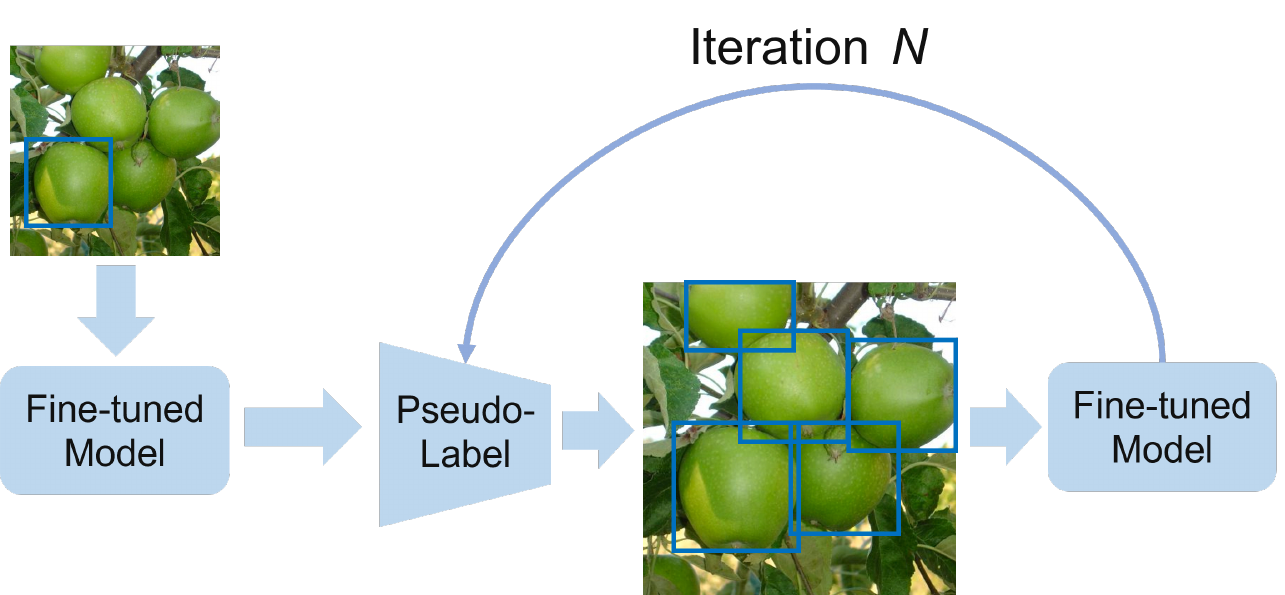} 
    \caption{Team IDCFS: overview of the iterative training process.}
    \label{fig:15_iter_train}
\end{figure}

\noindent \textbf{Iterative Training.} 
Given the scarcity, high cost, and limited availability of annotated data in few-shot learning scenarios, this team also designed an iterative training approach to train the model, as shown in Figure~\ref{fig:15_iter_train}. Specifically, the proposed method first fine-tunes the model for a few steps using the available labeled data. Then, the fine-tuned model is used to predict the support set samples, selecting the predictions with high confidence as pseudo-labels to update the label information of the support set samples. The model is then fine-tuned again. By iterating this process, the proposed method fully utilizes the information in the support set samples, achieving better performance while ensuring the robustness of the model, making it less susceptible to the influence of low-quality labels.

\noindent \textbf{Fine-tuning Grounding DINO with LoRA.} 
The IDCFS team also uses Grounding DINO~\cite{liu2024grounding} as another foundation model to generate bounding boxes and classification probabilities. The LoRA~\cite{hu2022lora} is used to fine-tune GroundingDINO on the few-shot training set. Specifically, this team adds bypass adapters to the linear projection layers (i.e., query, key, and value) of the attention mechanism in the visual backbone and BERT of Grounding DINO. 
To facilitate better adaptation to cross-domain datasets, the original model weights are frozen, and only the newly added parameters are trained.

\noindent \textbf{Model Ensemble.}
To effectively combine the outputs of GLIP and Grounding DINO, a model ensemble strategy with confidence reweighting is employed. Specifically, the detection scores from each model are scaled by pre-defined reliability weights. The reweighted predictions are then merged and refined using Non-Maximum Suppression (NMS)~\cite{neubeck2006efficient} to eliminate redundant bounding boxes and produce the final fused results. This approach allows the more reliable model to have a greater influence on the final predictions, enhancing detection performance by leveraging the complementary strengths of both models.

\subsubsection{Training Details}
For GLIP fine-tuning, the GLIP-L variant is used, which incorporates Swin-L~\cite{liu2021swin} as the visual encoder and BERT~\cite{devlin2019bert} as the text encoder. The model is pre-trained on a variety of datasets, including FourODs~\cite{kuznetsova2020open,krishna2017visual,krizhevsky2012imagenet}, GoldG~\cite{kamath2021mdetr}, CC3M+12M, and SBU~\cite{ordonez2011im2text}. During fine-tuning, full-model training is applied with a reduced learning rate of 2e-5, compared to the original setting of 1e-4 in GLIP.
For Grounding DINO, the Swin-B~\cite{liu2021swin} backbone is used as the visual encoder and BERT from Hugging Face~\cite{wolf2019huggingface} as the text encoder. The model is pre-trained on COCO~\cite{lin2014microsoft}, Objects365~\cite{shao2019objects365}, GoldG~\cite{kamath2021mdetr}, Cap4M, OpenImages~\cite{kuznetsova2020open}, ODinW-35~\cite{li2022elevater}, and RefCOCO~\cite{kamath2021mdetr}.
For the 1-shot and 5-shot settings on the CARPK dataset~\cite{hsieh2017drone}, no fine-tuning is performed. For 1-shot training on DeepFruits~\cite{sa2016deepfruits}, only the backbone is fine-tuned using LoRA. In all other cases, LoRA is used to fine-tune both the backbone and the BERT text encoder.

%% file: teams/team13_FDUROILab_Lenovo/main.tex
\subsection{FDUROILab$\_$Lenovo}
\subsubsection{Proposed Method}

\noindent\textbf{Efficient Tuning.}
To enhance the model’s adaptability in cross-domain few-shot detection (CDFSOD), this team proposes an efficient fine-tuning strategy. The proposed approach leverages data augmentation techniques to expand the training set and improve the model’s ability to recognize objects in the target domain with proposed k-shot annotated samples.

\begin{figure}[h]
    \centering
    \includegraphics[width=\linewidth]{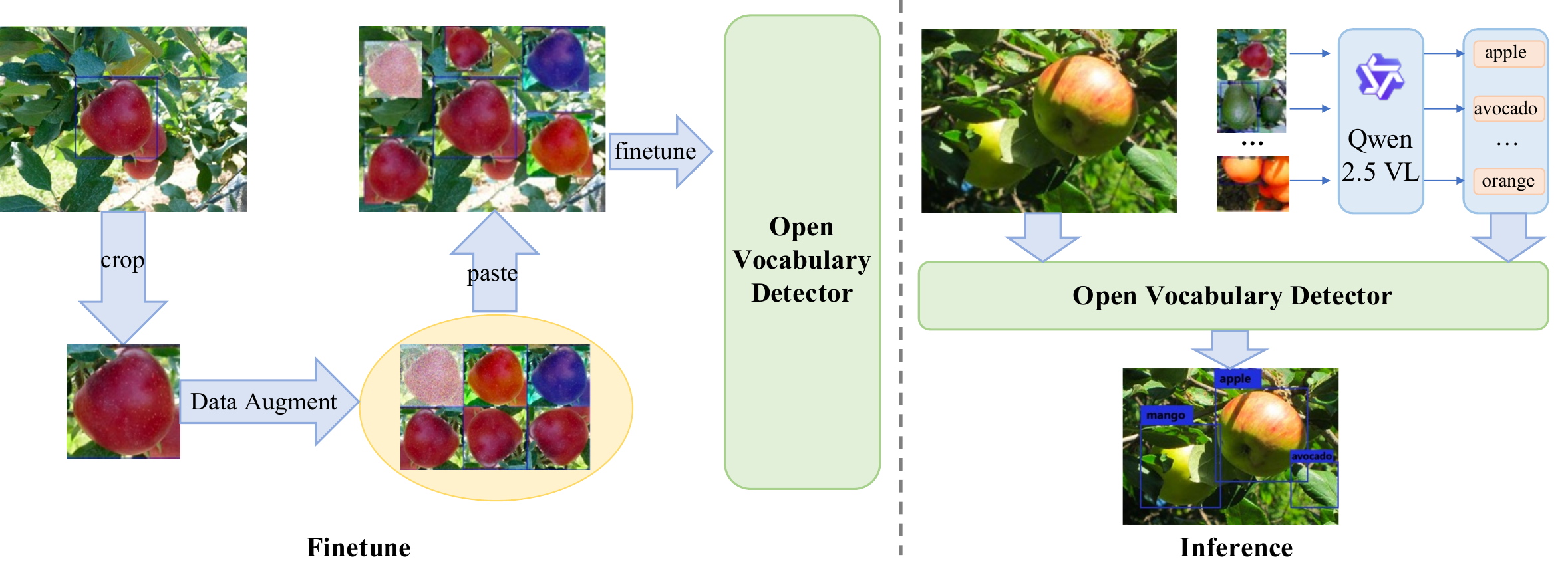} 
    \caption{Team FDUROILab$\_$Lenovo: overview of the efficient tuning and inference.}
    \label{fig:13_tuning}
\end{figure}

Specifically, given a k-shot setting, where k represents the number of provided object samples, the proposed approach adopts a structured fine-tuning pipeline, which is shown in Figure~\ref{fig:13_tuning}. 

(1) \textit{Object Cropping and Augmentation.} Using the provided bounding boxes of k-shot examples, the proposed method first crops the target objects from the original images. The cropped objects are then subjected to various data augmentation techniques, including flipping, rotation, grayscale conversion, and other transformations, to introduce diversity and improve generalization. (2) \textit{Object Rescaling and Random Pasting.} The proposed method randomly rescales the augmented objects to different sizes and pastes these transformed objects to the original images at different locations. This step simulates new object placements and enhances the model’s robustness to variations in object appearance and context. (3) \textit{Fine-Tuning with Augmented Data.} The proposed method finetunes the open-vocabulary detection model with the augmented images. This enables the detector to better adapt to objects in the target domain, even with minimal labeled examples. Additionally, the augmented data effectively increases the number of training samples, mitigating the few-shot learning limitation and improving overall detection performance. Through this efficient fine-tuning approach, the finetuned model gains enhanced adaptability to new domains while maintaining the advantages of open-vocabulary detection.

\noindent\textbf{Inference.} 
Since the proposed approach is based on an open-vocabulary detection model, it requires access to the target category labels during inference, which is shown in Figure~\ref{fig:13_tuning}. To obtain these labels, this team utilizes Qwen2.5-VL~\cite{bai2025qwen2} to generate the textual descriptions of the target categories. The retrieved target labels from Qwen2.5-VL are used as textual input to guide the detection process. Then, the open-vocabulary detection model~\cite{fu2025llmdet} is used to identify and classify objects in the test image based on the provided text-based labels.

\begin{figure}[h]
    \centering
    \includegraphics[width=1.\linewidth]{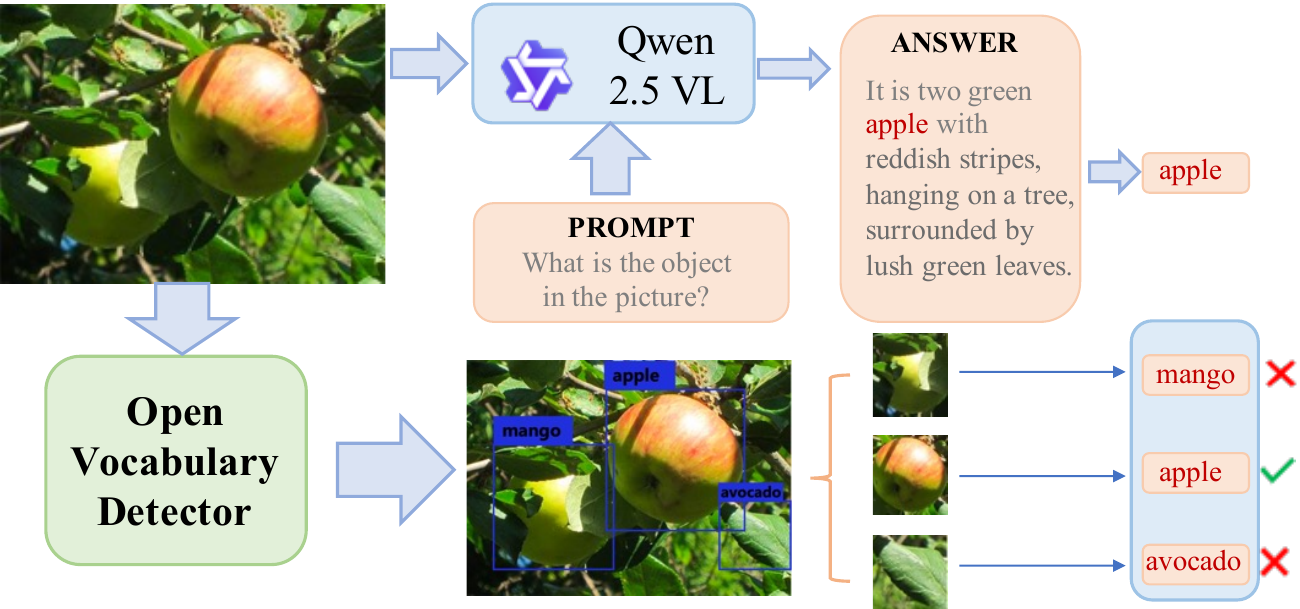} 
    \caption{Team FDUROILab$\_$Lenovo: post processing.}
    \label{fig:13_post}
\end{figure}

\noindent\textbf{Post-Process.} 
Although existing open-vocabulary detectors possess strong open-set detection capabilities, their performance on the challenge test set remains suboptimal. Upon further analysis, this team found that while the detector can successfully identify most objects, its primary weakness lies in classification errors rather than detection failures. This indicates that the open-vocabulary detection model still struggles with accurate classification when adapting to objects in a new domain. To address this issue, the Qwen2.5-VL is introduced as an auxiliary classifier to refine the final predictions, which is shown in Figure~\ref{fig:13_post}. For each test image, this team prompts Qwen2.5-VL to describe the objects present in the scene and provide a list of candidate categories that are likely to appear in the image. After that, this team refines the output of the open-vocabulary detection model using one of two strategies: (1) \textbf{Filtering.} Remove objects that are classified incorrectly by the detector and are not listed by Qwen2.5-VL. (2) \textbf{Reclassification}: Assign all detected objects to one of the categories predicted by Qwen2.5-VL, ensuring consistency between the detected bounding boxes and the high-level scene understanding of the multimodal model. The choice between these two strategies depends on the specific test dataset. 
By leveraging Qwen2.5-VL as a post-processing step, this team effectively corrects classification errors and enhances the model’s performance on unseen domains, leading to more accurate and reliable object detection results.

\subsubsection{Training Details}
LLMDet~\cite{fu2025llmdet} is adopted as the open-vocabulary detection model, with Swin-Large~\cite{liu2021swin} serving as the visual backbone. The Qwen2.5-VL-72B~\cite{bai2025qwen2} is introduced as the multimodal large language model (MLLM). Fine-tuning experiments are conducted on eight NVIDIA RTX 3090 GPUs, using a batch size of 8 and a learning rate of 1e-6. The number of training iterations varies across datasets and few-shot settings. For DeepFruits~\cite{sa2016deepfruits} and CarDD~\cite{wang2023cardd}, the model is fine-tuned for 30, 50, and 100 batches under the 1-shot, 5-shot, and 10-shot settings. No fine-tuning is performed for CARPK~\cite{hsieh2017drone}.

To enhance classification accuracy, dataset-specific post-processing strategies are applied. For DeepFruits, all detected objects are reclassified into one of the categories predicted by Qwen2.5-VL. In the case of CarDD, detected objects not belonging to the predefined categories are filtered out. As CARPK contains only a single object category, no additional classification is performed. However, further filtering is applied to remove overly large bounding boxes, which are likely to be incorrect, as the objects in this dataset are generally small. In all cases, Non-Maximum Suppression (NMS) is used to eliminate redundant or overlapping predictions.

%% file: teams/team14_HUSTLab/main.tex
\subsection{HUSTLab}
\subsubsection{Proposed Method}
The HUSTLab explores the usage of Qwen2.5VL, MM-GroundingDINO, and LLMDet for the open-source CD-FSOD. The proposed method can be divided into two distinct phases: 1) Obtaining text descriptions from the training set using the Qwen2.5VL model; 2) Selecting a base model, such as Grounding DINO or LLMDet, and fine-tuning it with CopyPaste data augmentation, followed by Adversarial Weight Perturbation (AWP) training to derive the final model and obtain test results. We observe that models like Grounding DINO possess robust object detection capabilities, and fine-tuning them with few-shot data significantly enhances detection performance in specific domains. Moreover, for training sets with limited samples, utilizing text descriptions generated by large-scale vision-language models proves highly effective.

\begin{figure}[h]
\centering	{\includegraphics[width=1.\linewidth]{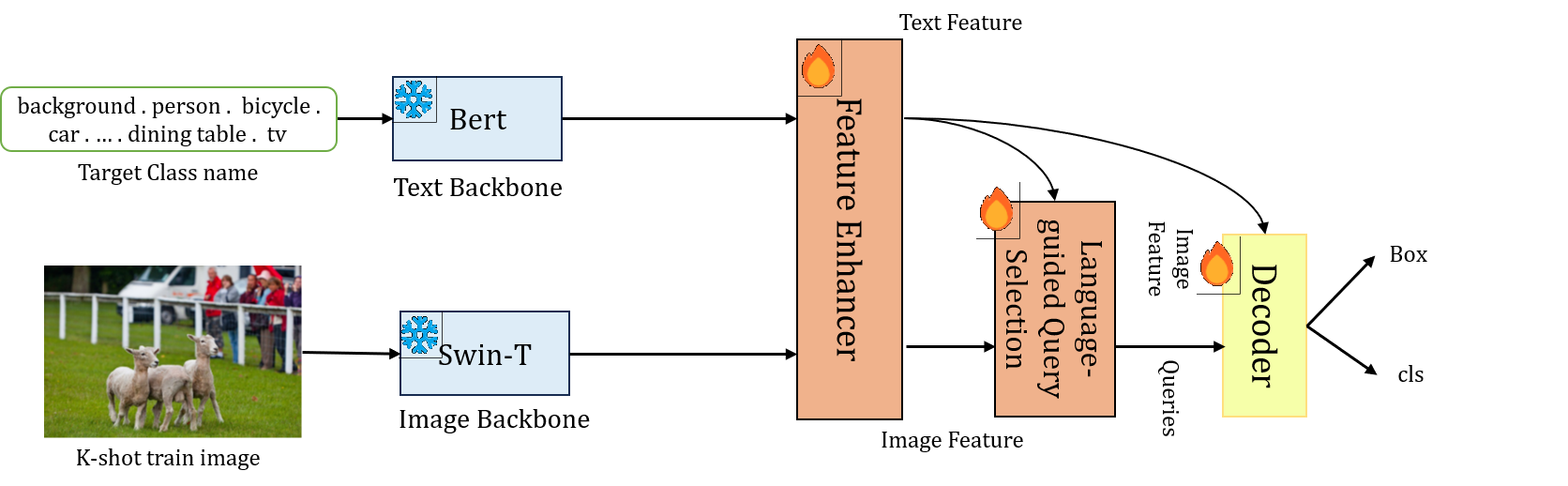}}
\caption{Team HUSTLab: overall framework of the proposed method. \label{fig:framework} 
}
\end{figure}

\noindent \textbf{Text Description Generation with a Large VLM.}
In this phase, this team leverages Qwen2.5VL to generate detailed text descriptions for the limited samples in the training set, extracting text-modal information from the images~\cite{pan2024solution}. Converting visual-modal information into text-modal information helps eliminate noise and condense semantic content. These detailed text descriptions are robust and will be fully utilized during the testing phase to enhance cross-domain few-shot object detection performance.

\begin{figure}[h]
\centering	{\includegraphics[width=1.\linewidth]{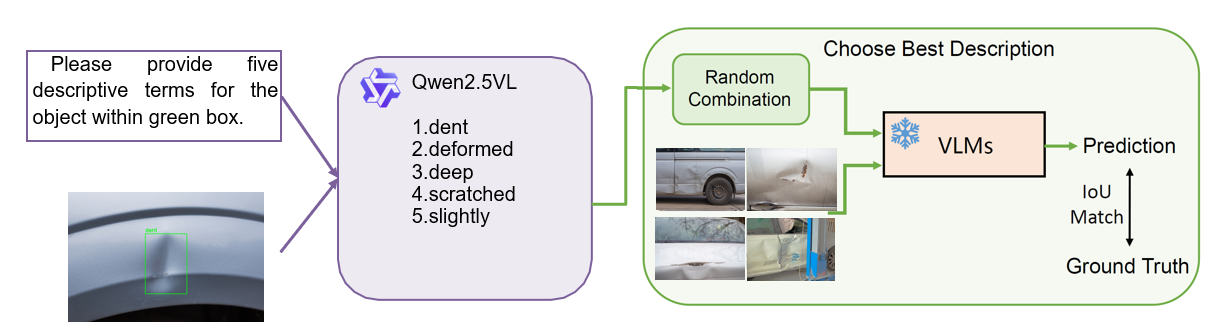}}
\caption{Team HUSTLab: text description generation~\cite{pan2024solution}. \label{fig:text description} 
}
\end{figure}

\noindent \textbf{Training Phase.}
In this stage, this team first selects an appropriate base model—either Grounding DINO\cite{liu2024grounding} or LLMDet—based\cite{fu2025llmdet} on its compatibility with the dataset. Using the zero-shot capabilities of the chosen base model, this team generates pseudo-labels, which are combined with ground-truth labels during training to regularize the model under few-shot conditions. To fine-tune the base model, this team uses CopyPaste\cite{ghiasi2021simple} data augmentation and Adversarial Weight Perturbation (AWP) techniques\cite{wu2020adversarial}. This approach strengthens the model’s generalization and robustness, enabling it to effectively handle cross-domain few-shot object detection tasks.

\subsubsection{Training Details}

The model is fine-tuned on three datasets using the MM-GroundingDINO-Large implementation provided by MMDetection as the base object detection framework, with the aim of enhancing cross-domain detection capabilities.
The performance largely depends on prompt design. Since part of the BERT-based text encoder is kept frozen during training, prompt quality plays a crucial role in boosting performance for certain object detection tasks. Prompts generated using Qwen2.5-VL are able to accurately describe the attribute features associated with abstract category names, thereby assisting the model in object localization and recognition. All experiments are conducted on 4 × NVIDIA RTX 3090 GPUs.

%% file: teams/team03_TongjiLab/main.tex
\subsection{TongjiLab}
\subsubsection{Proposed Method}
The TongjiLab proposes ProtoDINO, an innovative approach for CD-FSOD under the open-set setting, building upon GroundingDINO~\cite{liu2024grounding} as the baseline model. To improve the target classification performance of the baseline model, the CLIP model~\cite{radford2021learning, ilharco_gabriel_2021_5143773} is employed to extract both local and global image features from a limited set of target domain samples. These features are subsequently used to construct support sets, which serve as the foundation for building local prototype and global prototype networks, respectively. In addition, a text prototype network is developed using the CLIP model. During the target detection phase, visual features are extracted from each image query using CLIP. The L2 distances between these visual features and the local prototypes, global prototypes, and text prototypes are then computed, with these distances serving as one of the metrics for target classification. Furthermore, a car-damage-detection model\footnote{\href{https://huggingface.co/beingamit99/car_damage_detection/tree/main}{https://huggingface.co/beingamit99/car\_damage\_detection/tree/main}}, implemented as a vehicle appearance damage classification model based on the Vision Transformer (ViT), is incorporated. For the final target classification, matching probabilities derived from the GroundingDINO model, the car-damage-detection model, and the prototype networks~\cite{snell2017prototypical} are weighted and combined to produce the overall classification metric. 

\begin{figure}[h]
    \centering
    \includegraphics[width=1\linewidth]{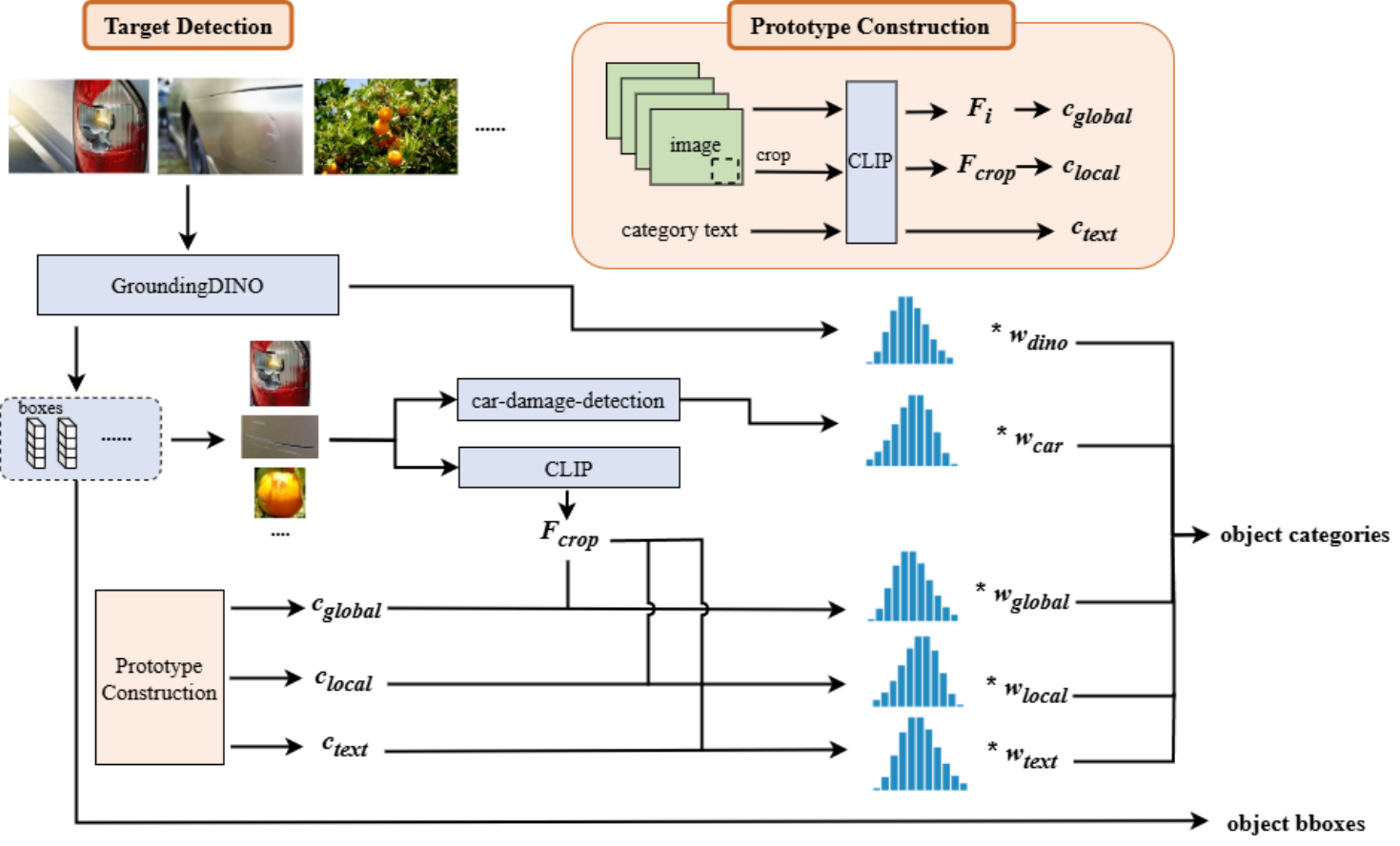}
    \caption{Team TongjiLab: framework of the proposed ProtoDINO.}
    \label{fig:ProtoDINO}
\end{figure}

The framework of the proposed ProtoDINO is depicted in Fig.~\ref{fig:ProtoDINO}. Overall, ProtoDINO operates in two key stages: prototype construction and target detection.

\noindent \textbf{Prototype Construction.}
During the prototype construction phase, this team crops few-shot learning images based on their annotations and generates visual embeddings as local feature prototypes $c_{local}$ for these local patches using the CLIP model. For 5-shot and 10-shot settings, $c_{local}$ is computed as the mean of all visual embeddings within the same category. Similarly, global feature prototypes $c_{global}$ are derived by encoding entire images through CLIP and applying the same averaging strategy across categories. For each category text $t$, this team builds the text prototype $c_{text}$ using CLIP as the text encoder.

\begin{equation}\label{eq:c_local}
c^{(n)}_{local} = \frac{1}{n}\sum_{i=1}^{n}F^{(i)}_{crop}
\end{equation}
\begin{equation}\label{eq:c_global}
c^{(n)}_{global} = \frac{1}{n}\sum_{i=1}^{n}F^{(i)}_{i}
\end{equation}
\begin{equation}\label{eq:c_local}
c^{(n)}_{text} = f_{clip\_text}(t^{(n)})
\end{equation}

\noindent \textbf{Target Detection.}
In the target detection stage, the input image and target category texts are processed by GroundingDINO to generate bounding boxes and initial classification probabilities. These bounding boxes are used to crop local regions from the image, which are then encoded by CLIP to obtain their visual features $F_{crop}$. 
To classify these regions, this team computes the L2 distances between their representations and the precomputed prototypes as in Eq.~\ref{eq:distance}. These distances are transformed into probability distributions via a softmax operation, yielding the prototype network’s classification output as in Eq.~\ref{eq:proto_prob}. Simultaneously, the cropped regions are evaluated by a pre-trained car-damage-detection model (based on Vision Transformer) to generate additional classification probabilities. The final classification decision is derived by aggregating probabilities from GroundingDINO, the car-damage-detection model, and the prototype network through a weighted summation as in Eq.~\ref{eq:final}. This fusion approach effectively integrates geometric localization from GroundingDINO, cross-modal semantics from CLIP, domain-specific insights from the car-damage-detection model, and few-shot prototype matching.

\begin{equation}\label{eq:distance}
d(u,v) = \sqrt[]{\sum_{n}(u^n-v^n)^2 } 
\end{equation}
\begin{equation}\label{eq:proto_prob}
probs_{proto} = -\frac{1}{\sigma}\cdot  e^{Norm[d(F,c)]} 
\end{equation}
\begin{equation}\label{eq:final}
probs = \sum_i w_i \cdot probs_i
\end{equation}

\subsubsection{Training Details}
The implementation is carried out on a server running CentOS 7, equipped with a single RTX 6000 Ada GPU. For the CLIP model, the DFN5B-CLIP-ViT-H-14-378 implementation is selected due to its balance between performance and efficiency in processing visual and textual data. For the GroundingDINO model, the official implementation is used. Based on empirical observations, the threshold parameter $\sigma$ is set to 0.5, which provides optimal results across various scenarios.
In GroundingDINO, the bounding box confidence threshold (\text{BOX\_THRESHOLD}) is set to 0.3. For the final decision fusion, the weighting coefficients for integrating outputs from multiple modules are empirically assigned as:
$ w_{\text{local}} = 0.25 $ (local prototype network),
$ w_{\text{global}} = 0.15 $ (global prototype network),
$ w_{\text{text}} = 0.4 $ (text prototype network),
$ w_{\text{dino}} = 0.1 $ (GroundingDINO), and
$ w_{\text{car}} = 0.1 $ (car-damage-detection model).

%% file: teams/team05_Manifold/main.tex
\subsection{Manifold}
\subsubsection{Proposed Method}
\begin{figure}[h]
    \vspace{-0.15in}
	\centering	{\includegraphics[width=1.\linewidth]{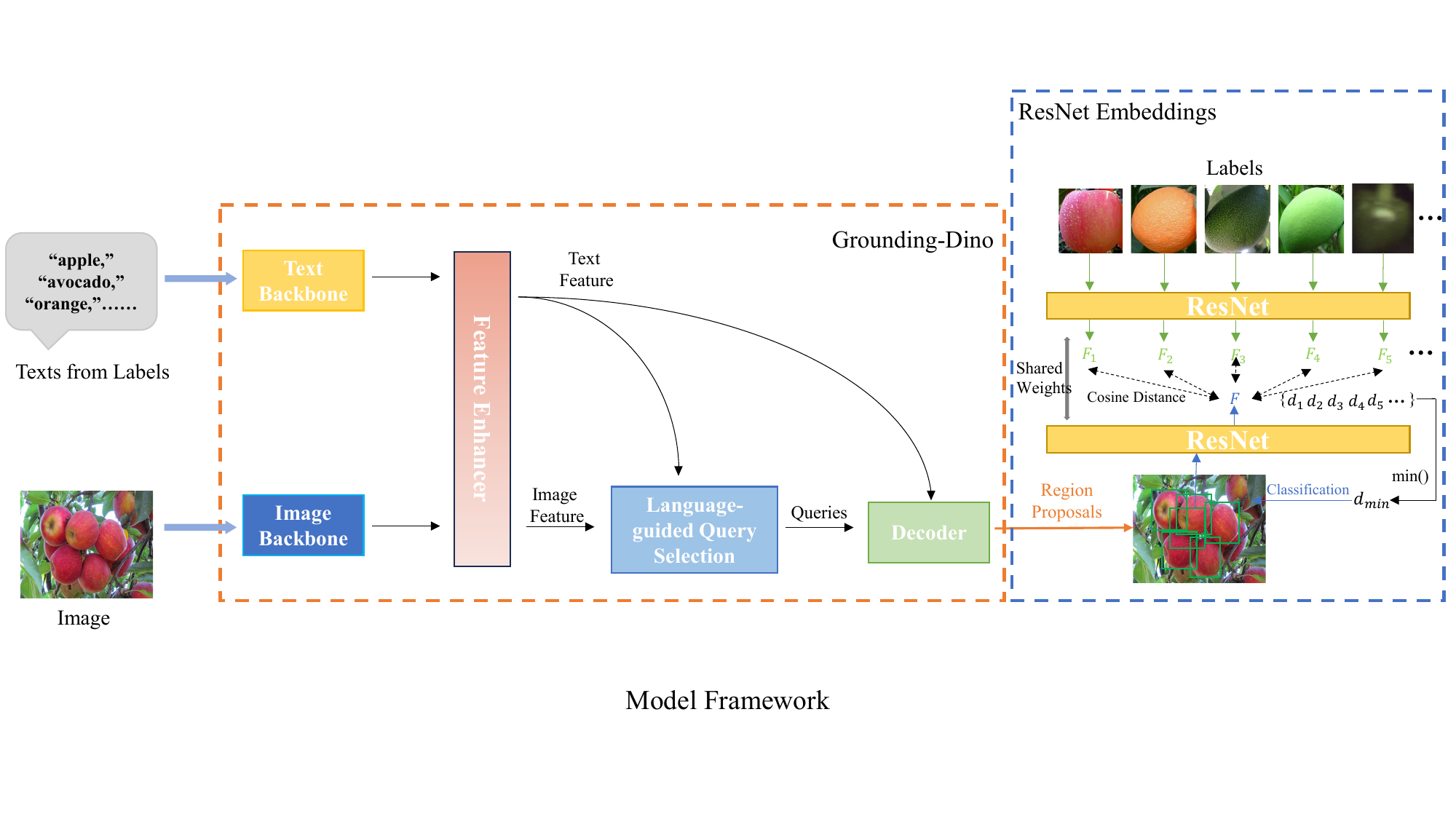}}
	\vspace{-0.35in}
	\caption{Team Manifold: overall framework of GDPRE. \label{gdpre_img} 
	}
\end{figure}

To address the challenge of few-shot object detection in cross-domain scenarios, the Manifold team proposes a novel approach based on the detection pipeline of a two-stage object detection algorithm. As illustrated in the Figure.~\ref{gdpre_img}, the proposed method first employs an open set object detection network, which is trained on public datasets, to detect objects in the query image. However, due to the domain gap between the pretraining datasets and the query datasets, the detection results cannot be directly trusted. Therefore, this team treats these results as region proposals that may contain objects of interest. Subsequently, this team combines the instance features from the support set for classification to obtain the final detection results.

\noindent \textbf{GroundingDINO-based Region Proposals.}
The GroundingDINO is selected as the pre-trained open-set object detector. It can detect objects of interest in images using input text, and it was pre-trained on seven datasets: COCO, O365, GoldG, Cap4M, OpenImage, ODinW-35, and RefCOCO. This pre-training gives it good detection capabilities for most real-world objects. However, in cross-domain few-shot scenarios, its detection effectiveness is suboptimal. For example, avocados may be misclassified as oranges because of the higher frequency of oranges in the pre-training data. Despite this, GroundingDINO can still provide region proposals for potential objects of interest in query images.

\noindent \textbf{ResNet-based Feature Classification.}
After obtaining region proposals, this team classifies the objects within them using support set images. Given the limited samples and significant intra-class variations in image space, directly matching support instances with query candidates in this space yields poor results. ResNet pre-trained on ImageNet is used to extract image features, mapping instances to a more robust feature space. To address scale differences, this team resizes instances in both support and region proposals images to 256×256 for feature extraction. Considering some classes have large intra-class and small inter-class differences, this team treats each instance's feature vector in multi-shot settings as a separate support vector rather than averaging them by class. This team calculates the cosine similarity between candidate region instances and support set instance feature vectors, assigning the region proposal instance to the class of the most similar support instance. This yields the final detection results, and the cosine similarity serves as the prediction confidence.

\subsubsection{Implementation Details}
Given that both GroundingDINO and ResNet are pre-trained on large-scale datasets, fine-tuning them under few-shot constraints—where the training classes do not overlap with the test classes—can be challenging. As a result, the pre-trained model weights are kept frozen. This approach requires minimal computational resources and can be executed on a laptop equipped with an RTX 4060 GPU. During inference, the category names from the test dataset are used as prompt inputs for GroundingDINO, and the BOX\_THRESHOLD is set to 0.1 to obtain the final detection results.

%% file: teams/team07_MXT/main.tex
\subsection{MXT}
\subsubsection{Proposed Method}
\begin{figure}[h]
	\centering	{\includegraphics[width=1.\linewidth]{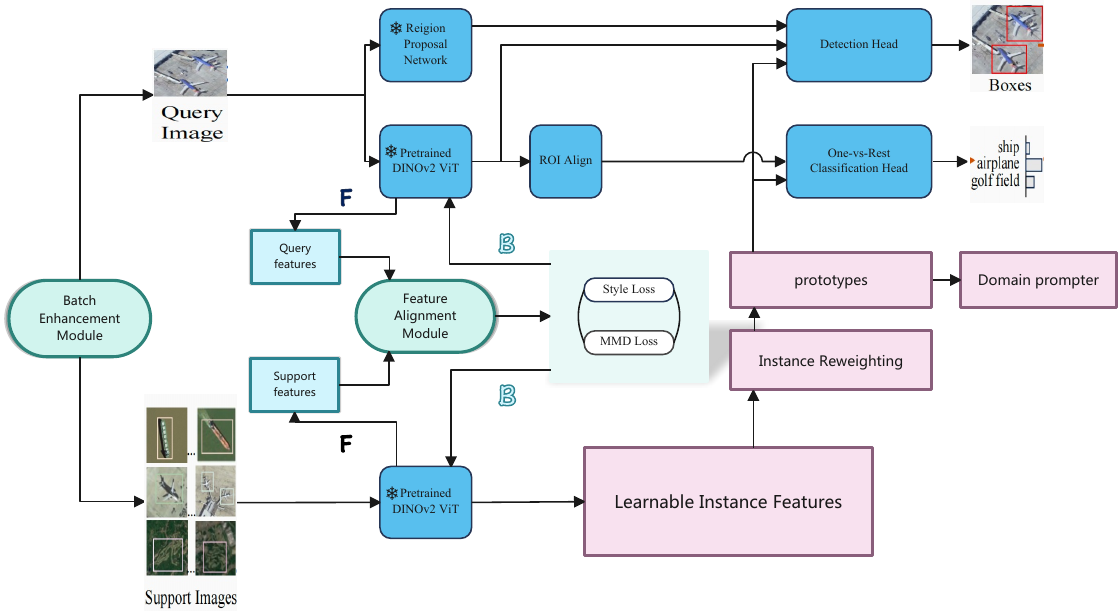}}
	\caption{Team DAEM: overall of the proposed model.}
	\label{mxt_img}
\end{figure}

This team proposes a Domain Adaptation Enhancement Module (DAEM) for Cross-Domain Few-Shot Object Detection (CD-FSOD), built as an extension to the CD-ViTO framework. While CD-ViTO provides a strong foundation for open-set cross-domain detection with DinoV2 ViT-L backbone, it still faces challenges with significant domain shifts. As illustrated in Fig~\ref{mxt_img}, the DAEM integrates seamlessly with the DinoV2 ViT-L backbone and enhances domain adaptation through two complementary mechanisms: batch enhancement and feature alignment.

\noindent \textbf{Batch Enhancement Module.}
The batch enhancement module increases training diversity through controlled style transfer between domains. For both source and target domain images, this team introduces cross-domain characteristics while preserving semantic content:

\begin{equation}
\text{img}_{styled} = \sigma_t \cdot \frac{\text{img} - \mu_s}{\sigma_s} + \mu_t
\end{equation}

where $\mu_s, \sigma_s$ are source image statistics and $\mu_t, \sigma_t$ are target domain statistics. The enhancement strength $\alpha$ gradually increases during training as follows:

\begin{equation}
\alpha = \min(1.0, \frac{t}{T_{warmup}})
\end{equation}

where $t$ is the current iteration and $T_{warmup}$ is set to 500. This gradual adaptation prevents disrupting the pre-trained DinoV2 ViT-L features early in training.

\noindent \textbf{Feature Alignment Module.}
The feature alignment module employs two complementary strategies to reduce domain gaps: Maximum Mean Discrepancy (MMD) and style-based adaptation.

\textit{MMD Loss}: 
The Maximum Mean Discrepancy is applied to reduce distribution differences between features from the source and target domains. MMD measures the distance between feature distributions in a reproducing kernel Hilbert space:

\begin{equation}
\mathcal{L}_{MMD}(\mathbf{X}_s, \mathbf{X}_t) = \left\|\frac{1}{n_s}\sum_{i=1}^{n_s}\phi(\mathbf{x}_s^i) - \frac{1}{n_t}\sum_{j=1}^{n_t}\phi(\mathbf{x}_t^j)\right\|_{\mathcal{H}}^2
\end{equation}

This is implemented with multiple Gaussian kernels with bandwidths $\sigma \in \{0.5, 1.0, 2.0, 5.0\}$ to capture similarities at different feature scales. This approach guides DinoV2 ViT-L to preserve its powerful representation abilities while adapting to target domains with minimal samples.

\textit{Style Loss}: Style-based adaptation addresses visual variations between domains that are unrelated to object semantics. For feature maps $\mathbf{F}$, the channel-wise statistics is transformed as:

\begin{equation}
\hat{\mathbf{F}} = \sigma_t \cdot \frac{\mathbf{F} - \mu_s}{\sigma_s} + \mu_t
\end{equation}

where $\mu_s, \sigma_s$ and $\mu_t, \sigma_t$ are the channel statistics of source and target features. This approach helps DinoV2 ViT-L focus on domain-invariant object characteristics rather than domain-specific visual styles.

The overall training objective combines the original CD-ViTO detection loss with the proposed domain adaptation components:

\begin{equation}
\mathcal{L} = \mathcal{L}_{det} + \lambda_{mmd}\mathcal{L}_{MMD} + \lambda_{style}\mathcal{L}_{style}
\end{equation}

\subsubsection{Training Details}
Following the pretrain–finetune–test pipeline established in the CD-FSOD benchmark, the pretrained DinoV2 ViT-L backbone from CD-ViTO is utilized. During fine-tuning, the backbone and Region Proposal Network (RPN) are selectively frozen, while the Domain-Adaptive Enhancement Modules (DAEM) and ROI Heads are optimized. This strategy preserves the general representational power of DinoV2 ViT-L while allowing domain-specific components to adapt effectively.

Training is conducted on NVIDIA A800 GPUs, with hyperparameters determined through extensive experimentation: the MMD loss weight is set to $\lambda_{mmd} = 0.16$, the style loss weight to $\lambda_{style} = 0.12$, and the batch enhancement strength to $\alpha_{max} = 0.8$. Differential learning rates are applied, using a multiplier of 2.0 for the DAEM modules and bias terms, with a base learning rate of $1 \times 10^{-4}$.

A warm-up phase of 500 iterations is introduced to gradually increase adaptation strength.
This helps stabilize early-stage training and prevents disruption of the pretrained DinoV2 ViT-L features. 
Optimization is performed using stochastic gradient descent (SGD) with a momentum of 0.9 and a weight decay of $1 \times 10^{-4}$. The model reaches optimal cross-domain performance after approximately 50 epochs. The proposed approach maintains the efficiency of CD-ViTO while delivering substantial improvements in challenging cross-domain few-shot detection scenarios.

%% file: teams/team12_X-Few/main.tex
\subsection{X-Few}
\subsubsection{Proposed Method}
\begin{figure}[h]
    \centering
    \includegraphics[width=1.\linewidth]{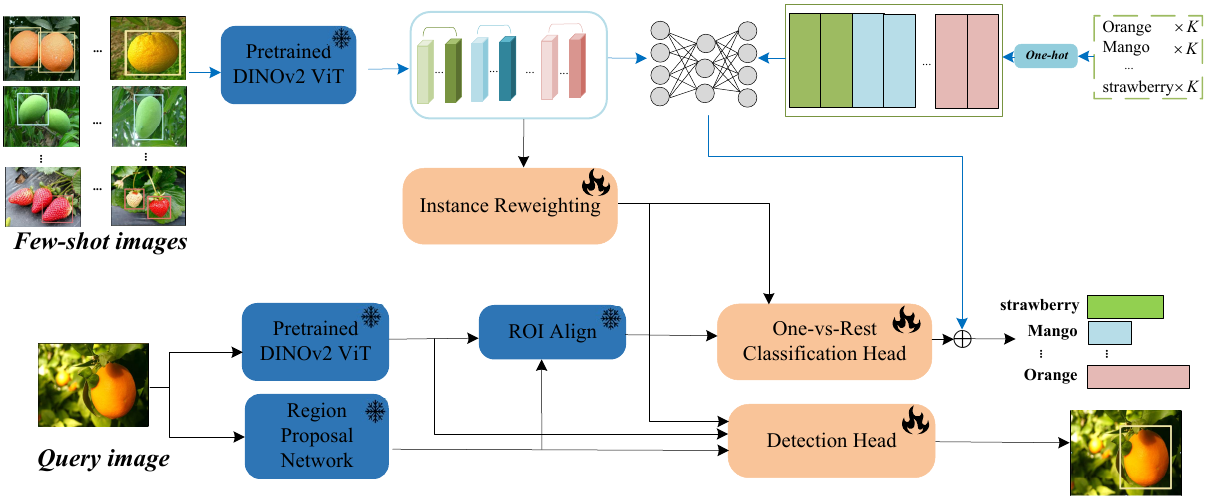}
    \caption{Team X-Few: illustration of the proposed Instance Feature Caching (IFC).}
    \label{fig:ifc-framework}
\end{figure}

To address the challenges of domain shift and category confusion arising from limited annotated data in CD-FSOD, the X-Few team proposes a novel domain adaptation strategy based on the Instance Feature Caching (IFC) mechanism.  The framework of the proposed method is shown in Fig.~\ref{fig:ifc-framework}, which is mainly built upon the CD-ViTO baseline.  Code is made available \footnote{\url{https://github.com/johnmaijer/X-Few-_CD-FSOD}}.

Intuitively, the IFC module is proposed to construct a cache model that could store and dynamically retrieve discriminative instance-level features from the target domain, alleviating model degradation caused by cross-domain distribution discrepancy in the few-shot supervision situation. Specifically, the IFC mechanism facilitates knowledge transfer through prototype-based feature alignment and an attention-guided memory update strategy, enhancing the model’s generalization capability in the data-scarce cross-domain scenario.

\textbf{Instance Feature Caching Construction.} 
Given a support set $\mathcal{S}$ comprising $N$ target categories, each consisting of $K$ annotated instances, denoted as $I_K$ with their associating labels $L_N$. For all $N \times K$ support samples, the proposed method leverages a pre-trained DINOv2 ViT $f_{CM}$ to obtain the instance-level features $F_{train} \in \mathbf{R}^{NK\times C}$. Similarly, the ground-truth labels are also encoded into \textit{N}-dimensional one-hot vectors $L_{train} \in \mathbf{R}^{NK \times N}$:
\begin{equation}
    F_{train} = \textbf{f}_{CM}(I_K)
\end{equation}
\begin{equation}
   L_{train} = \textbf{OneHot}(I_N)
\end{equation}
The feature extraction step is performed in an offline fashion to ensure persistent storage of high-quality feature representations for support set instances, thereby preserving discriminative semantic characteristics and spatial-aware contextual patterns in a memory-efficient manner. 
Then, these features and their corresponding label encodings are systematically cached to establish a comprehensive knowledge base that facilitates adaptive domain-aware detection while mitigating catastrophic forgetting.

\noindent\textbf{Instance Search.} After constructing the instance feature caching, given a query image $ \mathcal{L} $, the proposed method first feeds $ \mathcal{L} $  into both the Region Proposal Network and the Vision Transformer encoder to generate candidate regions and extract their deep features, respectively. These region proposals are then combined with the corresponding instance-level features in $ \mathcal{L} $  to derive a query vector $f_{test}$ for each candidate bounding box. 
Then, the proposed method achieves the most relevant instance feature lookup and finally calculate the adaptation representation $A \times L_{train}$ for the target domain, where $\textbf{A} \in \mathbf{R} ^{NK}$ is the affinity matrix between query vector and instance feature caching, being defined as:
\begin{equation}
    \textbf{A} = \exp(-\beta(1-f_{test}F_{train}^T))
\end{equation}
Ultimately, the domain adaptation representation is fed into the classification and regression branches of the original detection framework to calibrate prediction results from the open-set detector:
\begin{enumerate}
    \item \textbf{Classification Enhancement}: The similarity distribution between query features and cached features is leveraged to refine confidence estimates for the target domain categories through contrastive alignment.
    \item \textbf{Localization Refinement}: Retrieved instance localization priors are incorporated to constrain bounding box regression, thereby mitigating cross-domain localization biases caused by domain shifts.
\end{enumerate}
The above two strategies ensure that the detector adaptively aligns domain-invariant semantic representations while suppressing spurious correlations introduced by cross-domain discrepancies.

\subsubsection{Training Details}
A single RTX A800 GPU is used for the experiments.
The model is pre-trained on COCO and fine-tuned on novel support images. For the DeepFruit\cite{sa2016deepfruits}, Carpk\cite{hsieh2017drone}, and CarDD\cite{wang2023cardd}, the specific hyper-parameters settings are shown in the Tab.~\ref{tab:hyperparameters}. The tailored combination of learning rates and epoch schedules reflects a fine-grained tuning strategy to address domain heterogeneity across datasets, ensuring optimal trade-offs between generalization and task-specific optimization.

\begin{table}[h]
	\caption{Team X-Few: the hyper-parameters settings. }
	\centering
	\resizebox{0.5\textwidth}{!}{
		\begin{tabular}{c|ccc|ccc|ccc}
			\hline\hline\noalign{\smallskip}	
			\multirow{2}{*}{hyperparameter/shot} & \multicolumn{3}{c|}{DeepFruit~\cite{sa2016deepfruits}}  &\multicolumn{3}{c|}{Carpk~\cite{hsieh2017drone}} & \multicolumn{3}{c}{CarDD~\cite{wang2023cardd}} \\
			\cline{2-10}
			& 1&  5& 10 & 1&  5& 10 & 1& 5& 10 \\
			\hline\hline\noalign{\smallskip}
                Batch size 
			& 16 &  16 & 16 
			& 16 &  16 & 16 
			& 16 &  16 & 16  \\
			\hline
               Initial lr
			& 1e-3 &  1e-3 & 1e-3 
			& 1e-4 &  1e-4 & 1e-4 
			& 1e-3  & 1e-3  &1e-3   \\
			\hline
               Epoch
			& 40 &  100 & 200 
			& 40 &  80 &  100
			& 40 &  100 & 200   \\
			\hline
		\end{tabular}
	}
	\label{tab:hyperparameters}
\end{table}

%% file: teams/team08_MM/main.tex
\subsection{MM}
\subsubsection{Proposed Method}
\begin{figure}[h]
\centering	{\includegraphics[width=1.\linewidth]{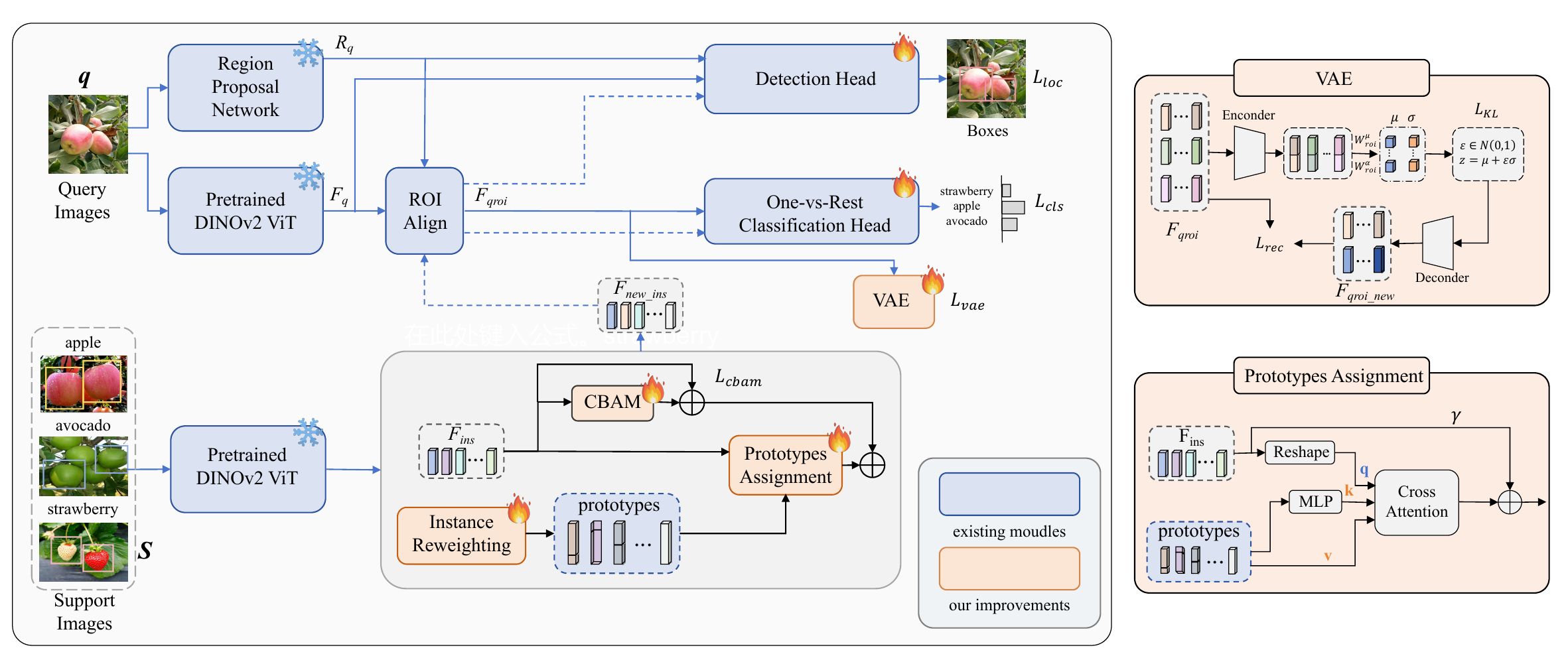}}
\caption{Team MM: overall framework of the DFE-ViT. 
}\label{fig:MMframework} 
\end{figure}

The MM team proposes a novel DFE-ViT method for CD-FSOD, in the closed set setting, which only takes COCO as the source data and transfers the model to a novel target. As in Fig.~\ref{fig:MMframework}, the proposed DFE-ViT method is built upon one open-set detector (DE-ViT) and finetuned using a few labeled instances from the target domain. New improvements include Instance Feature Enhancement, ROI Feature Enhancement. 

Specifically, given $S$ and $q$ as input, DFE-ViT follows a similar pipeline as DE-ViT to obtain instance features $F_{ins}$, region proposals $R_q$, visual features $F_q$, and ROI features $F_{q_{roi}}$. However, different from directly using $F_{ins}$ to derive the class prototypes, an  \textbf{Instance Feature Enhancement module (IFE)} and an \textbf{ROI Feature Enhancement module (RFE)} are proposed to enhance feature representation from both instance-level and ROI-level perspectives.

The \textbf{IFE} module adopts a residual CBAM structure to refine $F_{ins}^{ob}$, enabling the network to adaptively emphasize informative channels and spatial regions. To guide this attention process more explicitly, a dedicated CBAM loss $\mathcal{L}_{cbam}$ is designed, which encourages the enhanced instance features to align with salient regions in both spatial and channel dimensions. Furthermore, to enhance semantic alignment,  a class prototype enhancement mechanism is further incorporated where each object instance interacts with its corresponding class prototype via cross-attention, ensuring more discriminative and category-aware features. The output of IFE is optimized jointly with the standard detection losses, including the localization loss $\mathcal{L}_{loc}$, classification loss $\mathcal{L}_{cls}$, and the attention-guided loss $\mathcal{L}_{cbam}$.

For ROI features, this team introduces \textbf{RFE} based on a Variational Autoencoder (VAE). Each ROI feature $F_{q_{roi}}$ is encoded into a latent distribution and then reconstructed, which enables learning a more robust and expressive representation. A reconstruction loss $\mathcal{L}_{vae}$ is employed to ensure fidelity and consistency in the learned latent space. This ROI-level enhancement complements the instance-level refinement, offering a more diversified and generalized feature representation.

The top modules including the detection head $M_{DET}$ and the classification head $M_{CLS}$ are fine-tuned using the combined objective: 
\begin{equation}
\label{eq:mm7} 
\mathcal{L} = \mathcal{L}_{loc} + \mathcal{L}_{cls} + \alpha*\mathcal{L}_{cbam} + \beta*\mathcal{L}_{vae}.
\end{equation}
\noindent \textbf{Instance Feature Enhancement.}
The IFE module aims to refine instance features by integrating spatial/channel attention and semantic guidance. Given input instance features $F_{ins} \in \mathbb{R}^{B \times C \times H \times W}$, it first applies a residual CBAM to obtain spatially and channel-refined features $F_{cbam}$. 
Then, class prototypes $P \in \mathbb{R}^{N \times C}$ are used to semantically enhance the instance features via a cross-attention mechanism. Specifically, query and key projections are computed as $Q = W_q F_{ins}$ and $K = W_k P$, followed by attention: $A = \text{softmax}(QK^\top/\sqrt{d})$. The attended prototype features are added with a learnable weight $\gamma$, yielding $F_{proto}$. The final enhanced features are computed as $F_{enh} = F_{cbam} + F_{proto}$, which are more discriminative for downstream detection.

\noindent \textbf{ROI Feature Enhancement.}
The RFE module is based on a Variational Autoencoder and class prototype computation. As shown in Fig.~\ref{fig:MMframework}, the orange modules represent the newly proposed contributions: using VAE to model ROI features and enriching them with class prototypes. Given input ROI features $x \in \mathbb{R}^{N \times C \times k \times k}$, VAE encodes $x$ into latent mean $\mu \in \mathbb{R}^{N \times d}$ and log-variance $\log \sigma^2 \in \mathbb{R}^{N \times d}$ through linear layers. Latent variables are sampled as $z = \mu + \sigma \odot \epsilon$ using the reparameterization trick. Then, $z$ is decoded to reconstruct the ROI features $\hat{x} = \text{Decoder}(z)$. The reconstruction loss is computed as $L_{recon} = \frac{1}{N} \sum_{i=1}^{N} \| \hat{x}_i - x_i \|^2$, and the KL divergence loss regularizes the latent distribution: $L_{KL} = -\frac{1}{2} \sum_{i=1}^{N}(1 + \log \sigma_i^2 - \mu_i^2 - \sigma_i^2)$. The total VAE loss is $L_{vae} = L_{recon} + L_{KL}$. Finally, class prototypes are computed to further enhance feature representation across categories.

\subsubsection{Training Details}

The model is trained in the ``pretrain, finetune, and test" pipeline. 
Specifically, the base DE-ViT model pretrained on COCO is taken, then the $M_{DET}$, $M_{CLS}$, $IFE$ and $RFE$ are tuned on novel support images $S$ using the loss as in Eq.~\ref{eq:mm7}.
The hyperparameter $\alpha$ temperature for $\mathcal{L}_{cbam}$, $\beta$ temperature for $\mathcal{L}_{vae}$  are set as 0.3,0.4 for all the target datasets. While the value $N_{dom}$ means the number of virtual domains depending on the number of target classes $N$, specifically, $N_{dom} = 2*N$. 
The hyperparameter Top-K ($K$) in DE-ViT is set to 5. For datasets with the number of classes $N$ less than 5, $K$ is set to $N$. The trainable parameters are finetuned on 1-shot around 80 epochs, and on 5/10-shot around 50 epochs. The SGD with a learning rate of 0.002 is used as the optimizer. Experiments are performed on four A6000 GPUs.

%% file: teams/team16_FSV/main.tex
\subsection{FSV}
\subsubsection{Proposed Method}
\begin{figure}[h]
\centering	
\includegraphics[width=\linewidth]{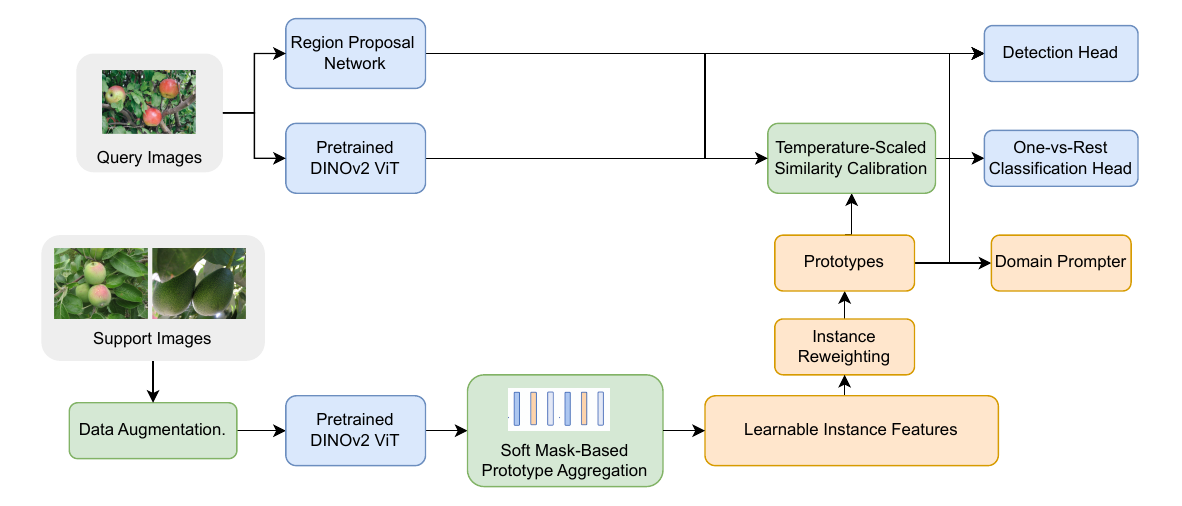}
\caption{Team FSV: overview of the proposed method.}
\vspace{-0.15in}
\label{fig:pipeline}
\end{figure}

The FSV team proposes an enhancement to the prototype-based detection for the cross-domain few-shot object detection (CD-FSOD) challenge under the closed-source setting, based on the CD-ViTO baseline model, as shown in Figure \ref{fig:pipeline}. 
Based on observations of the existing approach, this team found that CD-FSOD faces three key challenges. First, few-shot learning inherently suffers from limited example diversity. Second, conventional binary masking treats all spatial locations within an object region equally, which fails to prioritize more discriminative central regions over potentially noisy boundary areas. Third, standard cosine similarity calculations between query features and prototypes lack proper calibration, resulting in suboptimal separability across domain shifts. To solve these three challenges, this team explores three techniques: (1) Support Set Data Augmentation, (2) Soft Mask-Based Prototype Aggregation, and (3) Temperature-Scaled Similarity Calibration. 

\noindent\textbf{Support Set Data Augmentation.} 
For the support images, the proposed approach constructs a stochastic augmentation function to increase the diversity of the samples. 
DINOv2 \cite{oquab2023dinov2} is used as the feature extraction backbone for the augmented data, for its robust self-supervised learning capabilities and effective cross-domain transfer. The augmentation pipeline consists of a composition of transformations including \textit{Random Saturation, Random Contrast, Random Brightness, Random Flip, Random Rotation, Random Crop, Random Erasing}, and \textit{ Resize Shortest Edge}.

\noindent\textbf{Soft Mask-Based Prototype Aggregation.} 
To prioritize more discriminative central regions over potentially noisy boundary areas, the conventional binary masks are replaced by Gaussian soft masks to create soft spatial attention. 
Let $F_{ins} = \{F_{ins}^{ob}, F_{ins}^{bg}\}$ denote the extracted instance features and $M$ denote the binary mask of an instance. The soft mask could be defined $\tilde{M}$ as: $\tilde{M} = \frac{G_\sigma(M)}{\max G_\sigma(M)}$, where $G_\sigma$ is the Gaussian filter with standard deviation parameter $\sigma$. The extracted instance features for foreground objects $F_{ins}^{ob}$ are then weighted by the soft mask $\tilde{M}$, used as the initialization for learnable instance features. 

\noindent\textbf{Temperature-Scaled Similarity Calibration.} Finally, to calibrate image features to other domains, the proposed approach takes temperature scaling to make the final prototypes better match those in the new domain, which is a simple yet effective strategy to improve the discriminability of similarity scores. Let $F_{q_{roi}}$ denote the ROI features extracted from a query image using DINOv2. $F_{pro}$ denotes the prototype vector. The temperature scaling is applied during the cosine similarity computation as 
\begin{equation}
    s_{\tau} = \frac{F_{q_{roi}}^\top F_{pro}}{\tau \cdot \|F_{q_{roi}}\| \cdot \|F_{pro}\|},
\end{equation}
where $\tau$ is a temperature parameter that controls the sharpness of the similarity distribution. 
By tuning the temperature parameter, the entropy of the output distribution can be better modulated. 

\subsubsection{Implementation Details}
The training procedure utilizes only the provided few-shot datasets (1-shot, 5-shot, and 10-shot variants), without incorporating additional external data. The trainable parameters are finetuned for each testing dataset around 100 epochs. The training batch size is 16, with a base learning rate of 0.002. The parameter $\sigma$ in Soft Mask-Based Prototype Aggregation is set to 2.0. The parameter $\tau$ in Temperature-Scaled Similarity Calibration is set to 0.07. Experiments are performed on four NVIDIA A100 GPUs.

%% file: teams/team01_IPC/main.tex
\subsection{IPC}
\subsubsection{Proposed Method}
The IPC team utilizes CD-ViTO as the baseline, which is an improved version of the DE-ViT method, designed to enhance the cross-domain detection capability. To further mitigate performance degradation caused by cross-domain discrepancies and a very small number of test domain reference examples, this team was inspired by~\cite{ruan2024fully} to introduce a test-time adaptation algorithm during the inference phase. 

\begin{figure}[h]
    \centering
    \includegraphics[width=1\linewidth]{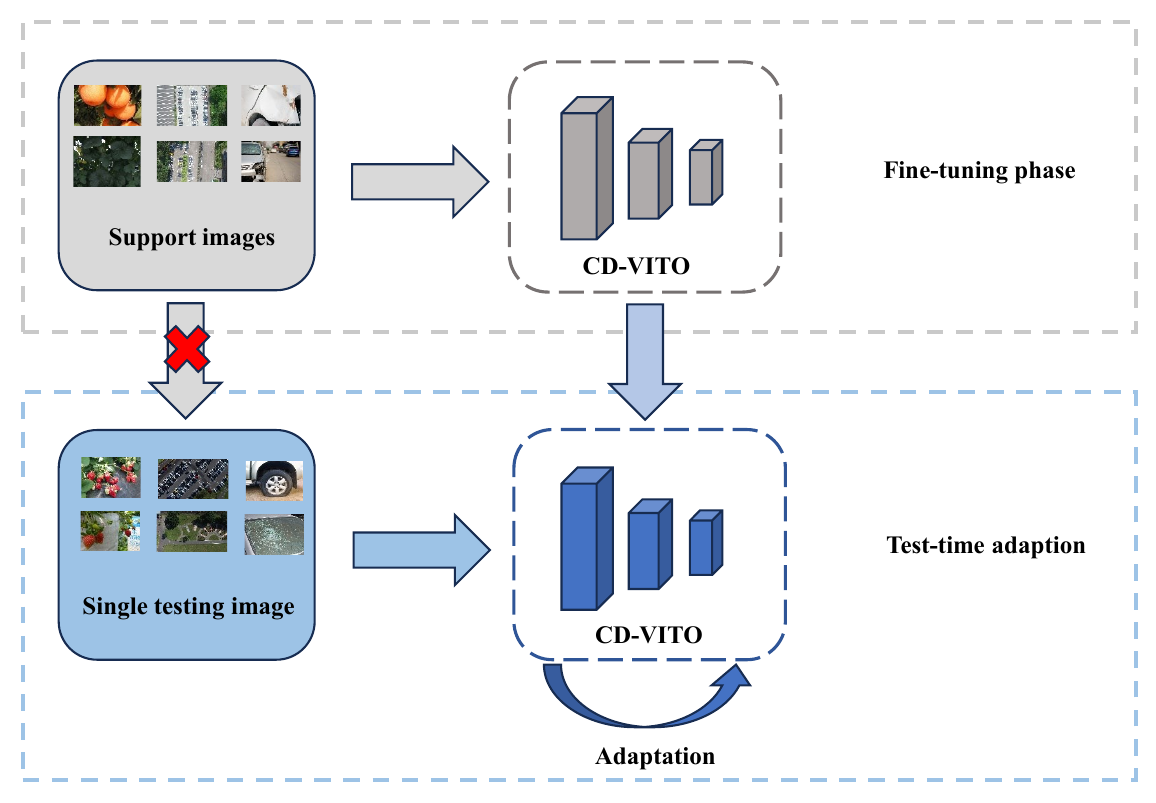}
    \caption{Team IPC: overview of the proposed approach. The upper section represents the baseline CD-ViTO fine-tuning phase; the lower section represents the test-time adaptation (TTA) process. The TTA procedure operates without access to the original training data, updating the fine-tuned detector on a single testing image before making a prediction. Crucially, only the mask prediction module in CD-ViTO undergoes gradient updates during TTA iterations.}
    \label{fig:tta}
\end{figure}

\begin{figure}[h]
    \centering
       \vspace{-0.15in}
    \includegraphics[width=1\linewidth]{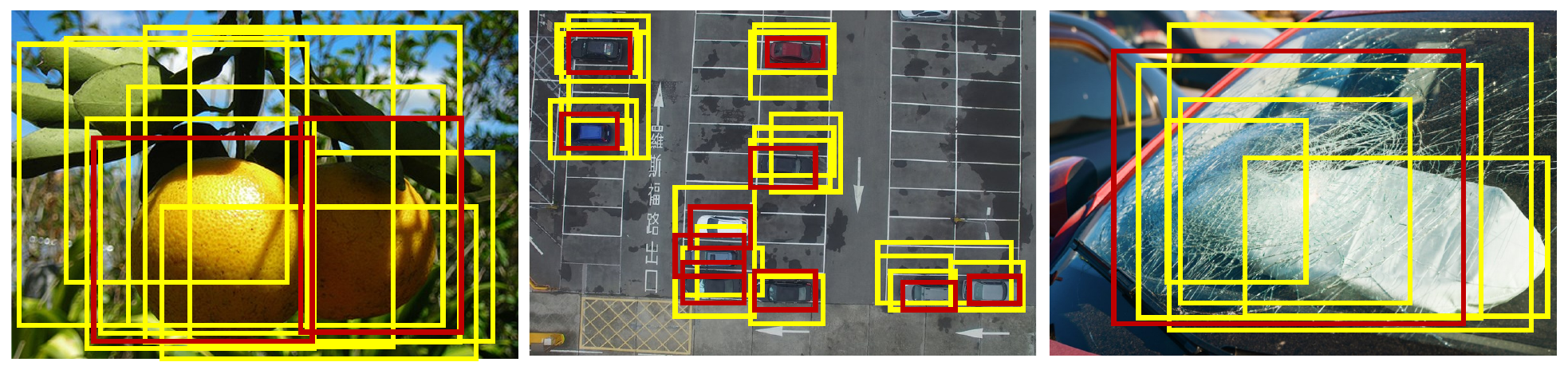}
    \vspace{-0.15in}
    \caption{Team IPC: by iteratively retaining proposals (yellow boxes \textcolor{yellow}{$\square$}) with high confidence scores as pseudo labels (red boxes \textcolor{red}{$\square$}), the model can effectively filter out most invalid detection boxes.}
    \label{fig:confidence_vis}
\end{figure}

To be specific, the proposed approach employs an iterative process as shown in Fig~\ref{fig:tta}. During each iteration $t$ (where $t \in \{1, \dots, T\}$), the existing detector $\theta_{t-1}$ generates predictions $D_t = \{(b_{t,i}, p_{t,i}) : \forall i\}$ for image $I$, with $b_{t,i}$ representing the $i^{th}$ object's bounding box and $p_{t,i} \in [0, 1]^K$ denoting the class probability distribution across $K$ categories. The detection confidence $c_{t,i} \in [0, 1]$ is determined by the highest probability in $p_{t,i}$, while the corresponding class index gives the predicted object category $y_{t,i} \in \{1, \dots, K\}$. Confident detections are then selected as pseudo-labels as illustrated in Fig~\ref{fig:confidence_vis}: $P_t = \{(b_{t,i}, y_{t,i}) : c_{t,i} > \lambda_{conf}\}$, with $\lambda_{conf}$ serving as the confidence cutoff. The detector is subsequently refined through gradient descent on these pseudo-labels, yielding an improved model $\theta_t$.

For the initial iteration ($t = 1$), the detector $\theta_{t-1}$ is initialized as $\theta_0$, which was pre-trained on source domain data. Upon completion of the final iteration ($t = T$), the optimized model $\theta_T$ produces the final predictions for $I$. Notably, this self-training paradigm maintains the original network architecture and operates without requiring access to source data or any other pretrained foundation models during adaptation.

\subsubsection{Training Details}
A single NVIDIA A6000 GPU is used for all experiments. The proposed method extends the CD-ViTO baseline through a test-time adaptation pipeline, initialized with k-shot instance fine-tuning on novel support datasets. During inference, the proposed method processes each test image using momentum SGD ($\beta=0.9, \alpha=0.001$) to exclusively update the mask prediction module through 5 iterations. For all experimental datasets, the cut-off confidence threshold $\lambda_{conf}$ is empirically set to 0.6.

%% file: teams/team11_LJY/main.tex
\subsection{LJY}
\def\Vec#1{{\boldsymbol{#1}}}
\def\Mat#1{{\boldsymbol{#1}}}
\subsubsection{Proposed Method}
\begin{figure}[ht]
    \centering
    \includegraphics[width=1.\linewidth]{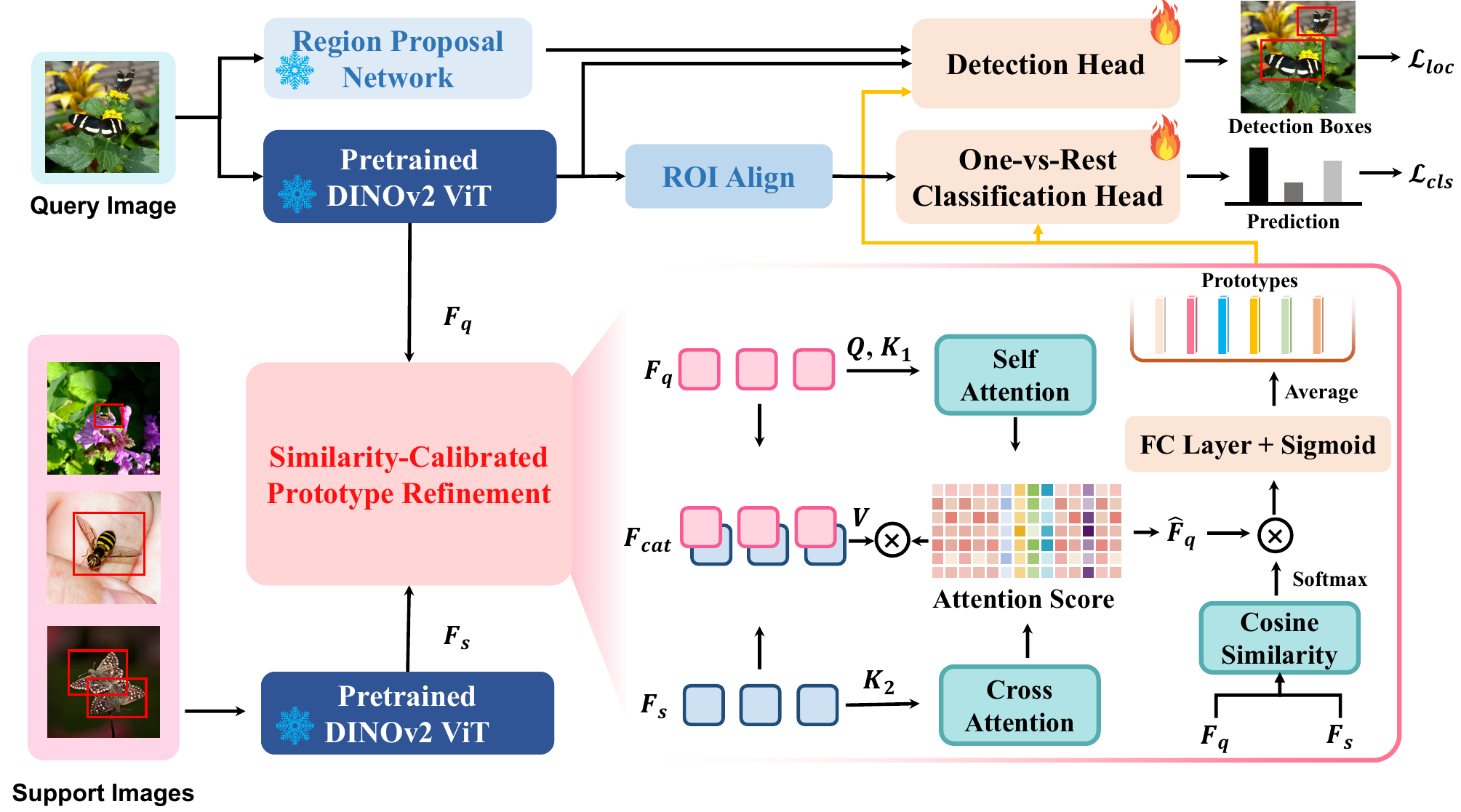}
    \caption{Team LJY: overall framework of SCPR.}
        \vspace{-0.1in}
    \label{fig:method}
\end{figure}

As shown in Fig.~\ref{fig:method}, the LJY team proposes similarity calibrated prototype refinement network, which utilizes query-aware guidelines to generate prototypes. The network contains a pretrained DINOv2 ViT, a region proposal network, an ROI align module, a detection head, and a one-vs-rest classification head. During the finetuning stage, the parameters of DINOv2 ViT are frozen. Only the parameters of the detection head and the classification head are finetuned.

Given a query image $\Mat{q} \in \mathbb{R}^{H \times W \times C}$ and a set of support images $S$, where $H$, $W$ and $C$ stand for the number of height, width and channels, respectively, the DINOv2 ViT backbone is used for obtaining query patches $\Vec{F}_q \in \mathbb{R}^{d}$ and support patches $\Vec{F}_s$.
Then, two linear layers are applied to project the query patches $\Vec{F}_q$ to $\Vec{Q}$ and $\Vec{K}_1$ and project the support patches $\Vec{F}_s$ to $\Vec{K}_2$.
The query patches $\Vec{F}_q$ and the support patches $\Vec{F}_s$ are then concatenated to obtain $\Vec{F}_{cat} = Concat(\Vec{F}_q, \Vec{F}_s)$. 
The concatenated patches $\Vec{F}_{cat}$ are projected to obtain $\Vec{V}$. To align the query patches and the support patches, the proposed method conducts scaled dot product on query patches $\Vec{F}_q$ and itself to obtain self attention score $ \Mat{A}_{self} = \frac{\Vec{Q} \Vec{K}_1^\top}{\sqrt{d}}$. Meanwhile, cross-attention score is computed using cosine similarity to ensure scale invariance $\Mat{A}_{cross} = \frac{\Vec{Q} \Vec{K}_2^\top}{||\Vec{Q}||_2||\Vec{K}_2||_2 + \epsilon}$
where $\epsilon$ is a small constant to avoid division by zero. The combined attention score is obtained by concatenating both and then be normalized by the softmax operation $\Mat{A} = Softmax(Concat(\Mat{A}_{self}, \Mat{A}_{cross}))$. The refined query representation is obtained by applying attention weights to the value matrix $\Vec{\hat{F}}_q = \Vec{F}_q + \Mat{A} \Vec{V}$. With the aligned query patches, the proposed method then generates prototypes with query-perceptual information. To further calibrate support features, their cosine similarity with the refined query is computed: $Sim = Softmax(\frac{\Vec{F}_s \Vec{F}_q^\top}{||\Vec{F}_s||_2||\Vec{F}_q||_2 + \epsilon})$. This similarity is used to re-weight the support representations: $\Vec{\hat{F}}_s = \Vec{F}_s + Sim * \Vec{\hat{F}}_q$. A learnable weighting function is applied via a sigmoid transformation: $\Vec{W} = Sigmoid(FC(\Vec{\hat{F}}_s))$. ensuring adaptive feature scaling: $\Vec{\hat{F}}_s = W \cdot \Vec{\hat{F}}_s$. The updated support features are then averaged across the K-shot dimension to derive refined prototypes:
$\Vec{P} = \frac{1}{K} \sum_{i=1}^K \Vec{\hat{F}}_s$. Finally, the query-aware prototype refinement is performed using a weighted combination of the refined prototypes and the original prototypes: $\Vec{\hat{P}} = \alpha \cdot \Vec{P} + (1-\alpha) \cdot \frac{1}{K} \sum_{i=1}^K \Vec{F}_s$. This final prototype representation retains both source-domain knowledge and query-specific adaptability, effectively enhancing cross-domain few-shot detection performance.

\subsubsection{Training Details}
The proposed modules are fine-tuned on novel support images, with the base DE-ViT pretrained on COCO  taken as initialization. The SGD with a learning rate of 0.002 is used as the optimizer. All experiments are conducted on two RTX3090 GPUs. The mAPs for 1/5/10 shots are reported.

%% file: teams/team10_MoveFree/affiliation.tex
\subsection*{MoveFree}
\noindent\textit{\textbf{Title:}} Marrying MoE-powered Grounding DINO with Self-training for Cross-domain Few-shot Object Detection\\
\noindent\textit{\textbf{Members: }} \\
Kaijin Zhang$^1$ (\href{zhang.kaijin1@zte.com.cn}{zhang.kaijin1@zte.com.cn}),\\
Qingpeng Nong$^1$ (\href{nong.qingpeng@zte.com.cn}{nong.qingpeng@zte.com.cn}),\\
Xiugang Dong$^1$ (\href{dong.xiugang20@zte.com.cn}{dong.xiugang20@zte.com.cn}),\\
Hong Gao$^1$ (\href{gao.hong@zte.com.cn}{gao.hong@zte.com.cn}),\\
Xiangsheng Zhou$^1$ (\href{zhou.xiangsheng@zte.com.cn}{zhou.xiangsheng@zte.com.cn})\\
\noindent\textit{\textbf{Affiliations: }} \\ 
$^1$ Central R \& D Institute, ZTE \\

%% file: teams/team06_AI4EarthLab/affiliation.tex
\subsection*{AI4EarthLab}
\noindent\textit{\textbf{Title:}} Enhance Then Search: An Augmentation-Search Strategy with Foundation Models for Cross-Domain Few-Shot Object Detection\\
\noindent\textit{\textbf{Members: }} \\
Jiancheng Pan$^1$ (\href{mailto:jiancheng.pan.plus@gmail.com}{jiancheng.pan.plus@gmail.com}),\\
Yanxing Liu$^2$(\href{mailto:liuyanxing21@mails.ucas.ac.cn}{liuyanxing21@mails.ucas.ac.cn}), \\
Xiao He$^3$(\href{mailto:xiaohewhu@163.com}{xiaohewhu@163.com}), \\
Jiahao Li$^1$(\href{mailto:lijiahao23@mails.tsinghua.edu.cn}{lijiahao23@mails.tsinghua.edu.cn}), \\
Yuze Sun$^1$(\href{mailto:syz23@mails.tsinghua.edu.cn}{syz23@mails.tsinghua.edu.cn}), \\
Xiaomeng Huang$^1$(\href{mailto:hxm@tsinghua.edu.cn}{hxm@tsinghua.edu.cn}) \\
\noindent\textit{\textbf{Affiliations: }} \\ 
$^1$ Tsinghua University \\
$^2$ University of Chinese Academy of Sciences \\
$^3$ Wuhan University \\

%% file: teams/team15_IDCFS/affiliation.tex
\subsection*{IDCFS}
\noindent\textit{\textbf{Title: }} Pseudo-Label Driven Vision-Language Grounding for Cross-Domain Few-Shot
Object Detection\\
\noindent\textit{\textbf{Members: }} \\
Zhenyu Zhang$^1$ (\href{mailto:yourEmail@xxx.xxx}{m202273680@hust.edu.cn}),\\
Ran Ma$^1$ (\href{mailto:yourEmail@xxx.xxx}{ranma@hust.edu.cn}),\\
Yuhan Liu$^1$ (\href{mailto:yourEmail@xxx.xxx}{yuhan\_liu@hust.edu.cn}),\\
Zijian Zhuang$^1$ (\href{mailto:yourEmail@xxx.xxx}{zhuangzj@hust.edu.cn}),\\
Shuai Yi$^1$ (\href{mailto:yourEmail@xxx.xxx}{yishuai@hust.edu.cn}),\\
Yixiong Zou$^1$ (\href{mailto:yourEmail@xxx.xxx}{yixiongz@hust.edu.cn})\\
\noindent\textit{\textbf{Affiliations: }} \\ 
$^1$ School of Computer Science and Technology, Huazhong University of Science and Technology \\

%% file: teams/team13_FDUROILab_Lenovo/affiliation.tex
\subsection*{FDUROILab\_Lenovo}
\noindent\textit{\textbf{Title: }} Efficient Tuning and MLLM-Based Post Prccessing for CDFSOD\\
\noindent\textit{\textbf{Members: }} \\
Lingyi Hong$^1$ (\href{lyhong22@m.fudan.edu.cn}{lyhong22@m.fudan.edu.cn}),\\
Mingxi Cheng$^1$(\href{mxchen24@m.fudan.edu.cn}{mxchen24@m.fudan.edu.cn}), \\
Runze Li$^2$(\href{lirz7@lenovo.com}{lirz7@lenovo.com}), \\
Xingdong Sheng$^2$(\href{shengxd1@lenovo.com}{shengxd1@lenovo.com}), \\
Wenqiang Zhang$^{1, 3}$(\href{wqzhang@fudan.edu.cn}{wqzhang@fudan.edu.cn}) \\
\noindent\textit{\textbf{Affiliations: }} \\ 
$^1$ Shanghai Key Lab of Intelligent Information Processing, School of Computer Science, Fudan University \\
$^2$ Lenovo Research \\
$^3$ Engineering Research Center of AI \& Robotics, Ministry of Education, Academy for Engineering \& Technology, Fudan University \\

%% file: teams/team14_HUSTLab/affiliation.tex
\subsection*{HUSTLab}
\noindent\textit{\textbf{Title: }} Prompt and Finetune Grounding DINO for Cross-Domain Few-shot Object Detection \\
\noindent\textit{\textbf{Members: }} \\
Weisen Chen$^1$ (\href{mailto:U202115027@hust.edu.cnx}{U202115027@hust.edu.cn}),\\
Yongxin Yan$^1$(\href{mailto:2585856499@qq.com}{2585856499@qq.com}),\\
Xinguo Chen$^2$(\href{mailto:327715@whut.edu.cn}{327715@whut.edu.cn}),\\
Yuanjie Shao$^1$(\href{mailto:shaoyuanjie@hust.edu.cn}{shaoyuanjie@hust.edu.cn}),\\
Zhengrong Zuo$^1$(\href{mailto:zhrzuo@main.hust.edu.cn}{zhrzuo@main.hust.edu.cn}),\\
Nong Sang$^1$(\href{mailto:nsang@hust.edu.cn}{nsang@hust.edu.cn})\\

\noindent\textit{\textbf{Affiliations: }} \\ 
$^1$ School of Artificial Intelligence and Automation, Huazhong University of Science and Technology \\
$^2$ School of Information Engineering,Wuhan University of Technology

%% file: teams/team03_TongjiLab/affiliation.tex
\subsection*{TongjiLab}
\noindent\textit{\textbf{Title: }}ProtoDINO: Cross-Domain Few-Shot Object Detection via GroundingDINO and CLIP-Based Prototypes\\
\noindent\textit{\textbf{Members: }} \\
Hao Wu$^1$ (\href{mailto:haowu@tongji.edu.cn}{haowu@tongji.edu.cn}),\\
Haoran Sun$^1$ \\
\noindent\textit{\textbf{Affiliations: }} \\ 
$^1$ Tongji University \\

%% file: teams/team05_Manifold/affiliation.tex
\subsection*{Manifold}
\noindent\textit{\textbf{Title: }} CDFSOD Challenge: Using Grounding-DINO Proposals and ResNet Embeddings\\
\noindent\textit{\textbf{Members: }} \\
Shuming Hu$^1$ (\href{mailto:hsm123@nudt.edu.cn}{hsm123@nudt.edu.cn}),\\
Yan Zhang$^1$, \\
Zhiguang Shi$^1$, \\
Yu Zhang$^1$, \\
Chao Chen$^1$,\\
Tao Wang$^1$\\
\noindent\textit{\textbf{Affiliations: }} \\ 
$^1$ National University of Defense Technology \\

%% file: teams/team07_MXT/affiliation.tex
\subsection*{MXT}
\noindent\textit{\textbf{Title: }}  Domain Adaptation Enhancement Module (DAEM) for Cross-Domain Few-Shot Object Detection \\
\noindent\textit{\textbf{Members: }} \\
Da Feng$^1$ (\href{mailto:yourEmail@xxx.xxx}{072108208@fzu.edu.cn}),\\
Linhai Zhuo$^1$ (\href{mailto:yourEmail@xxx.xxx}{534537916@qq.com}),\\
Ziming Lin$^1$ \\
\noindent\textit{\textbf{Affiliations: }} \\ 
$^1$ Fuzhou University \\

%% file: teams/team12_X-Few/affiliation.tex
\subsection*{X-Few}
\noindent\textit{\textbf{Title:}} IFC: Instance Feature Caching for Cross-Domain Few-Shot Object Detection\\
\noindent\textit{\textbf{Members: }} \\
Yali Huang$^1$ (\href{mailto:yourEmail@xxx.xxx}{hyl2024@gs.zzu.edu.cn}),\\
Jie Mei$^1$ (\href{mailto:yourEmail@xxx.xxx}{mj123123@gs.zzu.edu.cn}),\\
Yiming Yang$^1$ (\href{mailto:yourEmail@xxx.xxx}{yangyim637@gmail.com}),\\
Mi Guo$^1$ (\href{mailto:yourEmail@xxx.xxx}{mimi987836724@gs.zzu.edu.cn}),\\
Mingyuan Jiu$^{1,2,3}$ (\href{mailto:yourEmail@xxx.xxx}{iemyjiu@zzu.edu.cn}),\\
Mingliang Xu$^{1,2,3}$ (\href{mailto:yourEmail@xxx.xxx}{iexumingliang@zzu.edu.cn})\\
\noindent\textit{\textbf{Affiliations: }} \\ 
$^1$ School of Computer and Artificial Intelligence, Zhengzhou University \\ 
$^2$ Engineering Research Center of Intelligent Swarm Systems, Ministry of Education, Zhengzhou University \\
$^3$ National SuperComputing Center in Zhengzhou\\

%% file: teams/team08_MM/affiliation.tex
\subsection*{MM}
\noindent\textit{\textbf{Title: }} DFE-ViT: Dual Feature Enhancement Network for Cross-Domain Few-Shot Object Detection.\\
\noindent\textit{\textbf{Members: }} \\
Maomao Xiong$^1$ (\href{202314866@mail.sdu.edu.cn}{202314866@mail.sdu.edu.cn}),\\
Qunshu Zhang$^1$(\href{202414859@mail.sdu.edu.cn}{202414859@mail.sdu.edu.cn}), \\
Xinyu Cao$^1$(\href{202414842@mail.sdu.edu.cn}{202414842@mail.sdu.edu.cn}) \\
\noindent\textit{\textbf{Affiliations: }} \\ 
$^1$ Shandong University \\

%% file: teams/team16_FSV/affiliation.tex
\subsection*{FSV}
\noindent\textit{\textbf{Title: }} Enhanced Prototype-based Cross-domain Few-shot Object Detection\\
\noindent\textit{\textbf{Members: }} \\
Yuqing Yang$^1$ (\href{mailto:yyqyang101@gmail.com}{yyqyang101@gmail.com})\\
\noindent\textit{\textbf{Affiliations: }} \\ 
$^1$ George Mason University\\

%% file: teams/team01_IPC/affiliation.tex
\subsection*{IPC}
\noindent\textit{\textbf{Title: }} Test-time Adaptation Strategy for Cross-Domain Few-Shot Object Detection\\
\noindent\textit{\textbf{Members: }} \\
Dianmo Sheng$^1$ (\href{mailto:yourEmail@xxx.xxx}{dmsheng@mail.ustc.edu.cn}),\\
Xuanpu Zhao$^1$, \\
Zhiyu Li$^1$, \\
Xuyang Ding$^1$ \\
\noindent\textit{\textbf{Affiliations: }} \\ 
$^1$ University of Science and Technology of China \\

%% file: teams/team11_LJY/affiliation.tex
\subsection*{LJY}
\noindent\textit{\textbf{Title: }} Similarity-Calibrated Prototype Refinement for Cross-Domain Few-Shot Object Detection\\
\noindent\textit{\textbf{Members: }} \\
Wenqian Li (\href{wenqianli.li@seu.edu.cn}{wenqianli.li@seu.edu.cn})\\
\noindent\textit{\textbf{Affiliations: }} \\ 
Southeast University

%% file: main.bbl
\begin{thebibliography}{87}
\providecommand{\natexlab}[1]{#1}
\providecommand{\url}[1]{\texttt{#1}}
\expandafter\ifx\csname urlstyle\endcsname\relax
  \providecommand{\doi}[1]{doi: #1}\else
  \providecommand{\doi}{doi: \begingroup \urlstyle{rm}\Url}\fi

\bibitem[Bai et~al.(2025)Bai, Chen, Liu, Wang, Ge, Song, Dang, Wang, Wang, Tang, et~al.]{bai2025qwen2}
Shuai Bai, Keqin Chen, Xuejing Liu, Jialin Wang, Wenbin Ge, Sibo Song, Kai Dang, Peng Wang, Shijie Wang, Jun Tang, et~al.
\newblock Qwen2. 5-vl technical report.
\newblock \emph{arXiv preprint arXiv:2502.13923}, 2025.

\bibitem[Cai et~al.(2024)Cai, Jiang, Wang, Tang, Kim, and Huang]{cai2024survey}
Weilin Cai, Juyong Jiang, Fan Wang, Jing Tang, Sunghun Kim, and Jiayi Huang.
\newblock A survey on mixture of experts.
\newblock \emph{arXiv preprint arXiv:2407.06204}, 2024.

\bibitem[Chen et~al.(2025{\natexlab{a}})Chen, Liu, Gong, Wang, Sun, Wu, Timofte, Zhang, et~al.]{ntire2025srx4}
Zheng Chen, Kai Liu, Jue Gong, Jingkai Wang, Lei Sun, Zongwei Wu, Radu Timofte, Yulun Zhang, et~al.
\newblock Ntire 2025 challenge on image super-resolution (×4): Methods and results.
\newblock In \emph{Proceedings of the IEEE/CVF Conference on Computer Vision and Pattern Recognition (CVPR) Workshops}, 2025{\natexlab{a}}.

\bibitem[Chen et~al.(2025{\natexlab{b}})Chen, Wang, Liu, Gong, Sun, Wu, Timofte, Zhang, et~al.]{ntire2025face}
Zheng Chen, Jingkai Wang, Kai Liu, Jue Gong, Lei Sun, Zongwei Wu, Radu Timofte, Yulun Zhang, et~al.
\newblock Ntire 2025 challenge on real-world face restoration: Methods and results.
\newblock In \emph{Proceedings of the IEEE/CVF Conference on Computer Vision and Pattern Recognition (CVPR) Workshops}, 2025{\natexlab{b}}.

\bibitem[Conde et~al.(2025{\natexlab{a}})Conde, Timofte, et~al.]{ntire2025raw}
Marcos Conde, Radu Timofte, et~al.
\newblock Ntire 2025 challenge on raw image restoration and super-resolution.
\newblock In \emph{Proceedings of the IEEE/CVF Conference on Computer Vision and Pattern Recognition (CVPR) Workshops}, 2025{\natexlab{a}}.

\bibitem[Conde et~al.(2025{\natexlab{b}})Conde, Timofte, et~al.]{ntire2025rawrgb}
Marcos Conde, Radu Timofte, et~al.
\newblock Raw image reconstruction from {RGB} on smartphones. ntire 2025 challenge report.
\newblock In \emph{Proceedings of the IEEE/CVF Conference on Computer Vision and Pattern Recognition (CVPR) Workshops}, 2025{\natexlab{b}}.

\bibitem[Devlin et~al.(2019)Devlin, Chang, Lee, and Toutanova]{devlin2019bert}
Jacob Devlin, Ming-Wei Chang, Kenton Lee, and Kristina Toutanova.
\newblock Bert: Pre-training of deep bidirectional transformers for language understanding.
\newblock In \emph{Proceedings of the 2019 conference of the North American chapter of the association for computational linguistics: human language technologies, volume 1 (long and short papers)}, pages 4171--4186, 2019.

\bibitem[Drange(2019)]{GeirArTaxOr}
Geir Drange.
\newblock Arthropod taxonomy orders object detection dataset.
\newblock In \emph{https://doi.org/10.34740/kaggle/dsv/1240192}, 2019.

\bibitem[Ershov et~al.(2025)Ershov, Korchagin, Khalin, Panshin, Terekhin, Zaychenkova, Lobarev, Plokhotnyuk, Abramov, Zhdanov, Dorogova, Mamedov, Banic, Perevozchikov, Timofte, et~al.]{ntire2025night}
Egor Ershov, Sergey Korchagin, Alexei Khalin, Artyom Panshin, Arseniy Terekhin, Ekaterina Zaychenkova, Georgiy Lobarev, Vsevolod Plokhotnyuk, Denis Abramov, Elisey Zhdanov, Sofia Dorogova, Yasin Mamedov, Nikola Banic, Georgii Perevozchikov, Radu Timofte, et~al.
\newblock Ntire 2025 challenge on night photography rendering.
\newblock In \emph{Proceedings of the IEEE/CVF Conference on Computer Vision and Pattern Recognition (CVPR) Workshops}, 2025.

\bibitem[Fedus et~al.(2022)Fedus, Zoph, and Shazeer]{fedus2022switch}
William Fedus, Barret Zoph, and Noam Shazeer.
\newblock Switch transformers: Scaling to trillion parameter models with simple and efficient sparsity.
\newblock \emph{Journal of Machine Learning Research}, 23\penalty0 (120):\penalty0 1--39, 2022.

\bibitem[Fu et~al.(2025{\natexlab{a}})Fu, Yang, Mo, Yan, Wei, Meng, Xie, and Zheng]{fu2025llmdet}
Shenghao Fu, Qize Yang, Qijie Mo, Junkai Yan, Xihan Wei, Jingke Meng, Xiaohua Xie, and Wei-Shi Zheng.
\newblock Llmdet: Learning strong open-vocabulary object detectors under the supervision of large language models.
\newblock \emph{arXiv preprint arXiv:2501.18954}, 2025{\natexlab{a}}.

\bibitem[Fu et~al.(2021)Fu, Fu, and Jiang]{fu2021meta}
Yuqian Fu, Yanwei Fu, and Yu-Gang Jiang.
\newblock Meta-fdmixup: Cross-domain few-shot learning guided by labeled target data.
\newblock In \emph{Proceedings of the 29th ACM international conference on multimedia}, pages 5326--5334, 2021.

\bibitem[Fu et~al.(2022)Fu, Xie, Fu, Chen, and Jiang]{fu2022me}
Yuqian Fu, Yu Xie, Yanwei Fu, Jingjing Chen, and Yu-Gang Jiang.
\newblock Me-d2n: Multi-expert domain decompositional network for cross-domain few-shot learning.
\newblock In \emph{Proceedings of the 30th ACM international conference on multimedia}, pages 6609--6617, 2022.

\bibitem[Fu et~al.(2023)Fu, Xie, Fu, and Jiang]{fu2023styleadv}
Yuqian Fu, Yu Xie, Yanwei Fu, and Yu-Gang Jiang.
\newblock Styleadv: Meta style adversarial training for cross-domain few-shot learning.
\newblock In \emph{CVPR}, 2023.

\bibitem[Fu et~al.(2024)Fu, Wang, Pan, Huai, Qiu, Shangguan, Liu, Fu, Van~Gool, and Jiang]{fu2024cross}
Yuqian Fu, Yu Wang, Yixuan Pan, Lian Huai, Xingyu Qiu, Zeyu Shangguan, Tong Liu, Yanwei Fu, Luc Van~Gool, and Xingqun Jiang.
\newblock Cross-domain few-shot object detection via enhanced open-set object detector.
\newblock In \emph{European Conference on Computer Vision}, pages 247--264. Springer, 2024.

\bibitem[Fu et~al.(2025{\natexlab{b}})Fu, Qiu, Fu, Timofte, Sebe, Yang, Van~Gool, et~al.]{ntire2025cross}
Yuqian Fu, Xingyu Qiu, Bin Ren~Yanwei Fu, Radu Timofte, Nicu Sebe, Ming-Hsuan Yang, Luc Van~Gool, et~al.
\newblock Ntire 2025 challenge on cross-domain few-shot object detection: Methods and results.
\newblock In \emph{Proceedings of the IEEE/CVF Conference on Computer Vision and Pattern Recognition (CVPR) Workshops}, 2025{\natexlab{b}}.

\bibitem[Ghiasi et~al.(2021)Ghiasi, Cui, Srinivas, Qian, Lin, Cubuk, Le, and Zoph]{ghiasi2021simple}
Golnaz Ghiasi, Yin Cui, Aravind Srinivas, Rui Qian, Tsung-Yi Lin, Ekin~D Cubuk, Quoc~V Le, and Barret Zoph.
\newblock Simple copy-paste is a strong data augmentation method for instance segmentation.
\newblock In \emph{Proceedings of the IEEE/CVF conference on computer vision and pattern recognition}, pages 2918--2928, 2021.

\bibitem[Guo et~al.(2020)Guo, Codella, Karlinsky, Codella, Smith, Saenko, Rosing, and Feris]{guo2020broader}
Yunhui Guo, Noel~C Codella, Leonid Karlinsky, James~V Codella, John~R Smith, Kate Saenko, Tajana Rosing, and Rogerio Feris.
\newblock A broader study of cross-domain few-shot learning.
\newblock In \emph{Computer vision--ECCV 2020: 16th European conference, glasgow, UK, August 23--28, 2020, proceedings, part XXVII 16}, pages 124--141. Springer, 2020.

\bibitem[Han et~al.(2025)Han, Fan, Kong, Liao, Guo, Li, Timofte, et~al.]{ntire2025text}
Shuhao Han, Haotian Fan, Fangyuan Kong, Wenjie Liao, Chunle Guo, Chongyi Li, Radu Timofte, et~al.
\newblock Ntire 2025 challenge on text to image generation model quality assessment.
\newblock In \emph{Proceedings of the IEEE/CVF Conference on Computer Vision and Pattern Recognition (CVPR) Workshops}, 2025.

\bibitem[Hsieh et~al.(2017)Hsieh, Lin, and Hsu]{hsieh2017drone}
Meng-Ru Hsieh, Yen-Liang Lin, and Winston~H Hsu.
\newblock Drone-based object counting by spatially regularized regional proposal network.
\newblock In \emph{Proceedings of the IEEE international conference on computer vision}, pages 4145--4153, 2017.

\bibitem[Hu et~al.(2022)Hu, Shen, Wallis, Allen-Zhu, Li, Wang, Wang, Chen, et~al.]{hu2022lora}
Edward~J Hu, Yelong Shen, Phillip Wallis, Zeyuan Allen-Zhu, Yuanzhi Li, Shean Wang, Lu Wang, Weizhu Chen, et~al.
\newblock Lora: Low-rank adaptation of large language models.
\newblock \emph{ICLR}, 1\penalty0 (2):\penalty0 3, 2022.

\bibitem[Ilharco et~al.(2021)Ilharco, Wortsman, Wightman, Gordon, Carlini, Taori, Dave, Shankar, Namkoong, Miller, Hajishirzi, Farhadi, and Schmidt]{ilharco_gabriel_2021_5143773}
Gabriel Ilharco, Mitchell Wortsman, Ross Wightman, Cade Gordon, Nicholas Carlini, Rohan Taori, Achal Dave, Vaishaal Shankar, Hongseok Namkoong, John Miller, Hannaneh Hajishirzi, Ali Farhadi, and Ludwig Schmidt.
\newblock Openclip, 2021.

\bibitem[Inoue et~al.(2018)Inoue, Furuta, Yamasaki, and Aizawa]{inoue2018cross}
Naoto Inoue, Ryosuke Furuta, Toshihiko Yamasaki, and Kiyoharu Aizawa.
\newblock Cross-domain weakly-supervised object detection through progressive domain adaptation.
\newblock In \emph{CVPR}, 2018.

\bibitem[Jain et~al.(2025)Jain, Wu, Zou, Florentin, Turbell, Siddhartha, Timofte, et~al.]{ntire2025vqe}
Varun Jain, Zongwei Wu, Quan Zou, Louis Florentin, Henrik Turbell, Sandeep Siddhartha, Radu Timofte, et~al.
\newblock Ntire 2025 challenge on video quality enhancement for video conferencing: Datasets, methods and results.
\newblock In \emph{Proceedings of the IEEE/CVF Conference on Computer Vision and Pattern Recognition (CVPR) Workshops}, 2025.

\bibitem[Jiang et~al.(2024)Jiang, Sablayrolles, Roux, Mensch, Savary, Bamford, Chaplot, Casas, Hanna, Bressand, et~al.]{jiang2024mixtral}
Albert~Q Jiang, Alexandre Sablayrolles, Antoine Roux, Arthur Mensch, Blanche Savary, Chris Bamford, Devendra~Singh Chaplot, Diego de~las Casas, Emma~Bou Hanna, Florian Bressand, et~al.
\newblock Mixtral of experts.
\newblock \emph{arXiv preprint arXiv:2401.04088}, 2024.

\bibitem[Jiang et~al.(2021)Jiang, Wang, Jia, Xu, Liu, Fan, Li, Liu, Xue, and Wang]{jiang2021underwater}
Lihao Jiang, Yi Wang, Qi Jia, Shengwei Xu, Yu Liu, Xin Fan, Haojie Li, Risheng Liu, Xinwei Xue, and Ruili Wang.
\newblock Underwater species detection using channel sharpening attention.
\newblock In \emph{ACM MM}, 2021.

\bibitem[Kamath et~al.(2021)Kamath, Singh, LeCun, Synnaeve, Misra, and Carion]{kamath2021mdetr}
Aishwarya Kamath, Mannat Singh, Yann LeCun, Gabriel Synnaeve, Ishan Misra, and Nicolas Carion.
\newblock Mdetr-modulated detection for end-to-end multi-modal understanding.
\newblock In \emph{Proceedings of the IEEE/CVF international conference on computer vision}, pages 1780--1790, 2021.

\bibitem[K{\"o}hler et~al.(2023)K{\"o}hler, Eisenbach, and Gross]{kohler2023few}
Mona K{\"o}hler, Markus Eisenbach, and Horst-Michael Gross.
\newblock Few-shot object detection: A comprehensive survey.
\newblock \emph{IEEE Transactions on Neural Networks and Learning Systems}, 2023.

\bibitem[Krishna et~al.(2017)Krishna, Zhu, Groth, Johnson, Hata, Kravitz, Chen, Kalantidis, Li, Shamma, et~al.]{krishna2017visual}
Ranjay Krishna, Yuke Zhu, Oliver Groth, Justin Johnson, Kenji Hata, Joshua Kravitz, Stephanie Chen, Yannis Kalantidis, Li-Jia Li, David~A Shamma, et~al.
\newblock Visual genome: Connecting language and vision using crowdsourced dense image annotations.
\newblock \emph{International journal of computer vision}, 123:\penalty0 32--73, 2017.

\bibitem[Krizhevsky et~al.(2012)Krizhevsky, Sutskever, and Hinton]{krizhevsky2012imagenet}
Alex Krizhevsky, Ilya Sutskever, and Geoffrey~E Hinton.
\newblock Imagenet classification with deep convolutional neural networks.
\newblock In \emph{Advances in Neural Information Processing Systems}, pages 1097--1105, 2012.

\bibitem[Kuznetsova et~al.(2020)Kuznetsova, Rom, Alldrin, Uijlings, Krasin, Pont-Tuset, Kamali, Popov, Malloci, Kolesnikov, et~al.]{kuznetsova2020open}
Alina Kuznetsova, Hassan Rom, Neil Alldrin, Jasper Uijlings, Ivan Krasin, Jordi Pont-Tuset, Shahab Kamali, Stefan Popov, Matteo Malloci, Alexander Kolesnikov, et~al.
\newblock The open images dataset v4: Unified image classification, object detection, and visual relationship detection at scale.
\newblock \emph{International journal of computer vision}, 128\penalty0 (7):\penalty0 1956--1981, 2020.

\bibitem[Lee et~al.(2025)Lee, Park, Canelo, Park, Kim, Chun, Jin, Li, Guo, Timofte, et~al.]{ntire2025ebhdr}
Sangmin Lee, Eunpil Park, Angel Canelo, Hyunhee Park, Youngjo Kim, Hyungju Chun, Xin Jin, Chongyi Li, Chun-Le Guo, Radu Timofte, et~al.
\newblock Ntire 2025 challenge on efficient burst hdr and restoration: Datasets, methods, and results.
\newblock In \emph{Proceedings of the IEEE/CVF Conference on Computer Vision and Pattern Recognition (CVPR) Workshops}, 2025.

\bibitem[Li et~al.(2022{\natexlab{a}})Li, Liu, Li, Zhang, Aneja, Yang, Jin, Hu, Liu, Lee, et~al.]{li2022elevater}
Chunyuan Li, Haotian Liu, Liunian Li, Pengchuan Zhang, Jyoti Aneja, Jianwei Yang, Ping Jin, Houdong Hu, Zicheng Liu, Yong~Jae Lee, et~al.
\newblock Elevater: A benchmark and toolkit for evaluating language-augmented visual models.
\newblock \emph{Advances in Neural Information Processing Systems}, 35:\penalty0 9287--9301, 2022{\natexlab{a}}.

\bibitem[Li et~al.(2020)Li, Wan, Cheng, Meng, and Han]{li2020object}
Ke Li, Gang Wan, Gong Cheng, Liqiu Meng, and Junwei Han.
\newblock Object detection in optical remote sensing images: A survey and a new benchmark.
\newblock \emph{ISPRS}, 2020.

\bibitem[Li et~al.(2022{\natexlab{b}})Li, Zhang, Zhang, Yang, Li, Zhong, Wang, Yuan, Zhang, Hwang, et~al.]{li2022grounded}
Liunian~Harold Li, Pengchuan Zhang, Haotian Zhang, Jianwei Yang, Chunyuan Li, Yiwu Zhong, Lijuan Wang, Lu Yuan, Lei Zhang, Jenq-Neng Hwang, et~al.
\newblock Grounded language-image pre-training.
\newblock In \emph{Proceedings of the IEEE/CVF conference on computer vision and pattern recognition}, pages 10965--10975, 2022{\natexlab{b}}.

\bibitem[Li et~al.(2022{\natexlab{c}})Li, Liu, and Bilen]{li2022cross}
Wei-Hong Li, Xialei Liu, and Hakan Bilen.
\newblock Cross-domain few-shot learning with task-specific adapters.
\newblock In \emph{Proceedings of the IEEE/CVF conference on computer vision and pattern recognition}, pages 7161--7170, 2022{\natexlab{c}}.

\bibitem[Li et~al.(2025{\natexlab{a}})Li, Jin, Jin, Wu, Li, Wang, Yang, Li, Chen, Wen, Tan, Timofte, et~al.]{ntire2025day}
Xin Li, Yeying Jin, Xin Jin, Zongwei Wu, Bingchen Li, Yufei Wang, Wenhan Yang, Yu Li, Zhibo Chen, Bihan Wen, Robby Tan, Radu Timofte, et~al.
\newblock Ntire 2025 challenge on day and night raindrop removal for dual-focused images: Methods and results.
\newblock In \emph{Proceedings of the IEEE/CVF Conference on Computer Vision and Pattern Recognition (CVPR) Workshops}, 2025{\natexlab{a}}.

\bibitem[Li et~al.(2025{\natexlab{b}})Li, Wang, Li, Yuan, Shao, Yao, Sun, Zhou, Timofte, and Chen]{ntire2025shortugc_data}
Xin Li, Xijun Wang, Bingchen Li, Kun Yuan, Yizhen Shao, Suhang Yao, Ming Sun, Chao Zhou, Radu Timofte, and Zhibo Chen.
\newblock Ntire 2025 challenge on short-form ugc video quality assessment and enhancement: Kwaisr dataset and study.
\newblock In \emph{Proceedings of the IEEE/CVF Conference on Computer Vision and Pattern Recognition (CVPR) Workshops}, 2025{\natexlab{b}}.

\bibitem[Li et~al.(2025{\natexlab{c}})Li, Yuan, Li, Guan, Shao, Yu, Wang, Lu, Luo, Yao, Sun, Zhou, Chen, Timofte, et~al.]{ntire2025shortugc}
Xin Li, Kun Yuan, Bingchen Li, Fengbin Guan, Yizhen Shao, Zihao Yu, Xijun Wang, Yiting Lu, Wei Luo, Suhang Yao, Ming Sun, Chao Zhou, Zhibo Chen, Radu Timofte, et~al.
\newblock Ntire 2025 challenge on short-form ugc video quality assessment and enhancement: Methods and results.
\newblock In \emph{Proceedings of the IEEE/CVF Conference on Computer Vision and Pattern Recognition (CVPR) Workshops}, 2025{\natexlab{c}}.

\bibitem[Liang et~al.(2025)Liang, Timofte, Yi, Zhang, Liu, Sun, Wu, Zhang, Zeng, Zhang, et~al.]{ntire2025raim}
Jie Liang, Radu Timofte, Qiaosi Yi, Zhengqiang Zhang, Shuaizheng Liu, Lingchen Sun, Rongyuan Wu, Xindong Zhang, Hui Zeng, Lei Zhang, et~al.
\newblock Ntire 2025 the 2nd restore any image model {(RAIM)} in the wild challenge.
\newblock In \emph{Proceedings of the IEEE/CVF Conference on Computer Vision and Pattern Recognition (CVPR) Workshops}, 2025.

\bibitem[Lin et~al.(2014)Lin, Maire, Belongie, Hays, Perona, Ramanan, Doll{\'a}r, and Zitnick]{lin2014microsoft}
Tsung-Yi Lin, Michael Maire, Serge Belongie, James Hays, Pietro Perona, Deva Ramanan, Piotr Doll{\'a}r, and C~Lawrence Zitnick.
\newblock Microsoft coco: Common objects in context.
\newblock In \emph{Computer vision--ECCV 2014: 13th European conference, zurich, Switzerland, September 6-12, 2014, proceedings, part v 13}, pages 740--755. Springer, 2014.

\bibitem[Liu et~al.(2024{\natexlab{a}})Liu, Feng, Wang, Wang, Liu, Zhao, Dengr, Ruan, Dai, Guo, et~al.]{liu2024deepseek}
Aixin Liu, Bei Feng, Bin Wang, Bingxuan Wang, Bo Liu, Chenggang Zhao, Chengqi Dengr, Chong Ruan, Damai Dai, Daya Guo, et~al.
\newblock Deepseek-v2: A strong, economical, and efficient mixture-of-experts language model.
\newblock \emph{arXiv preprint arXiv:2405.04434}, 2024{\natexlab{a}}.

\bibitem[Liu et~al.(2024{\natexlab{b}})Liu, Zeng, Ren, Li, Zhang, Yang, Jiang, Li, Yang, Su, et~al.]{liu2024grounding}
Shilong Liu, Zhaoyang Zeng, Tianhe Ren, Feng Li, Hao Zhang, Jie Yang, Qing Jiang, Chunyuan Li, Jianwei Yang, Hang Su, et~al.
\newblock Grounding dino: Marrying dino with grounded pre-training for open-set object detection.
\newblock In \emph{European Conference on Computer Vision}, pages 38--55. Springer, 2024{\natexlab{b}}.

\bibitem[Liu et~al.(2025{\natexlab{a}})Liu, Min, Hu, Zhang, Guo, et~al.]{ntire2025xgc}
Xiaohong Liu, Xiongkuo Min, Qiang Hu, Xiaoyun Zhang, Jie Guo, et~al.
\newblock Ntire 2025 {XGC} quality assessment challenge: Methods and results.
\newblock In \emph{Proceedings of the IEEE/CVF Conference on Computer Vision and Pattern Recognition (CVPR) Workshops}, 2025{\natexlab{a}}.

\bibitem[Liu et~al.(2025{\natexlab{b}})Liu, Wu, Vasluianu, Yan, Ren, Zhang, Gu, Zhang, Zhu, Timofte, et~al.]{ntire2025lowlight}
Xiaoning Liu, Zongwei Wu, Florin-Alexandru Vasluianu, Hailong Yan, Bin Ren, Yulun Zhang, Shuhang Gu, Le Zhang, Ce Zhu, Radu Timofte, et~al.
\newblock Ntire 2025 challenge on low light image enhancement: Methods and results.
\newblock In \emph{Proceedings of the IEEE/CVF Conference on Computer Vision and Pattern Recognition (CVPR) Workshops}, 2025{\natexlab{b}}.

\bibitem[Liu et~al.(2021)Liu, Lin, Cao, Hu, Wei, Zhang, Lin, and Guo]{liu2021swin}
Ze Liu, Yutong Lin, Yue Cao, Han Hu, Yixuan Wei, Zheng Zhang, Stephen Lin, and Baining Guo.
\newblock Swin transformer: Hierarchical vision transformer using shifted windows.
\newblock In \emph{Proceedings of the IEEE/CVF International Conference on Computer Vision}, pages 10012--10022, 2021.

\bibitem[Neubeck and Van~Gool(2006)]{neubeck2006efficient}
Alexander Neubeck and Luc Van~Gool.
\newblock Efficient non-maximum suppression.
\newblock In \emph{18th international conference on pattern recognition (ICPR'06)}, pages 850--855. IEEE, 2006.

\bibitem[Oquab et~al.(2023)Oquab, Darcet, Moutakanni, Vo, Szafraniec, Khalidov, Fernandez, Haziza, Massa, El-Nouby, et~al.]{oquab2023dinov2}
Maxime Oquab, Timoth{\'e}e Darcet, Th{\'e}o Moutakanni, Huy Vo, Marc Szafraniec, Vasil Khalidov, Pierre Fernandez, Daniel Haziza, Francisco Massa, Alaaeldin El-Nouby, et~al.
\newblock Dinov2: Learning robust visual features without supervision.
\newblock \emph{arXiv preprint arXiv:2304.07193}, 2023.

\bibitem[Ordonez et~al.(2011)Ordonez, Kulkarni, and Berg]{ordonez2011im2text}
Vicente Ordonez, Girish Kulkarni, and Tamara Berg.
\newblock Im2text: Describing images using 1 million captioned photographs.
\newblock \emph{Advances in neural information processing systems}, 24, 2011.

\bibitem[Pan et~al.(2024{\natexlab{a}})Pan, Yi, Yang, Qi, Hu, Xu, and Yang]{pan2024solution}
Hongpeng Pan, Shifeng Yi, Shouwei Yang, Lei Qi, Bing Hu, Yi Xu, and Yang Yang.
\newblock The solution for cvpr2024 foundational few-shot object detection challenge.
\newblock \emph{arXiv preprint arXiv:2406.12225}, 2024{\natexlab{a}}.

\bibitem[Pan et~al.(2024{\natexlab{b}})Pan, Liu, Fu, Ma, Li, Paudel, Gool, and Huang]{pan2024locateearthadvancingopenvocabulary}
Jiancheng Pan, Yanxing Liu, Yuqian Fu, Muyuan Ma, Jiaohao Li, Danda~Pani Paudel, Luc~Van Gool, and Xiaomeng Huang.
\newblock Locate anything on earth: Advancing open-vocabulary object detection for remote sensing community, 2024{\natexlab{b}}.

\bibitem[Pan et~al.(2024{\natexlab{c}})Pan, Ma, Ma, Bai, and Chen]{pan2024pir}
Jiancheng Pan, Muyuan Ma, Qing Ma, Cong Bai, and Shengyong Chen.
\newblock Pir: Remote sensing image-text retrieval with prior instruction representation learning, 2024{\natexlab{c}}.

\bibitem[Qiao et~al.(2021)Qiao, Zhao, Li, Qiu, Wu, and Zhang]{qiao2021defrcn}
Limeng Qiao, Yuxuan Zhao, Zhiyuan Li, Xi Qiu, Jianan Wu, and Chi Zhang.
\newblock Defrcn: Decoupled faster r-cnn for few-shot object detection.
\newblock In \emph{ICCV}, 2021.

\bibitem[Radford et~al.(2021)Radford, Kim, Hallacy, Ramesh, Goh, Agarwal, Sastry, Askell, Mishkin, Clark, et~al.]{radford2021learning}
Alec Radford, Jong~Wook Kim, Chris Hallacy, Aditya Ramesh, Gabriel Goh, Sandhini Agarwal, Girish Sastry, Amanda Askell, Pamela Mishkin, Jack Clark, et~al.
\newblock Learning transferable visual models from natural language supervision.
\newblock In \emph{International conference on machine learning}, pages 8748--8763. PmLR, 2021.

\bibitem[Ren et~al.(2023)Ren, Liu, Song, Bi, Cucchiara, Sebe, and Wang]{ren2023masked}
Bin Ren, Yahui Liu, Yue Song, Wei Bi, Rita Cucchiara, Nicu Sebe, and Wei Wang.
\newblock Masked jigsaw puzzle: A versatile position embedding for vision transformers.
\newblock In \emph{Proceedings of the IEEE/CVF Conference on Computer Vision and Pattern Recognition}, pages 20382--20391, 2023.

\bibitem[Ren et~al.(2024{\natexlab{a}})Ren, Li, Liang, Ranjan, Liu, Cucchiara, Gool, Yang, and Sebe]{ren2024sharing}
Bin Ren, Yawei Li, Jingyun Liang, Rakesh Ranjan, Mengyuan Liu, Rita Cucchiara, Luc~V Gool, Ming-Hsuan Yang, and Nicu Sebe.
\newblock Sharing key semantics in transformer makes efficient image restoration.
\newblock \emph{Advances in Neural Information Processing Systems}, 37:\penalty0 7427--7463, 2024{\natexlab{a}}.

\bibitem[Ren et~al.(2025)Ren, Guo, Sun, Wu, Timofte, Li, et~al.]{ntire2025esr}
Bin Ren, Hang Guo, Lei Sun, Zongwei Wu, Radu Timofte, Yawei Li, et~al.
\newblock The tenth ntire 2025 efficient super-resolution challenge report.
\newblock In \emph{Proceedings of the IEEE/CVF Conference on Computer Vision and Pattern Recognition (CVPR) Workshops}, 2025.

\bibitem[Ren et~al.(2024{\natexlab{b}})Ren, Jiang, Liu, Zeng, Liu, Gao, Huang, Ma, Jiang, Chen, et~al.]{ren2024grounding}
Tianhe Ren, Qing Jiang, Shilong Liu, Zhaoyang Zeng, Wenlong Liu, Han Gao, Hongjie Huang, Zhengyu Ma, Xiaoke Jiang, Yihao Chen, et~al.
\newblock Grounding dino 1.5: Advance the" edge" of open-set object detection.
\newblock \emph{arXiv preprint arXiv:2405.10300}, 2024{\natexlab{b}}.

\bibitem[Ruan and Tang(2024)]{ruan2024fully}
Xiaoqian Ruan and Wei Tang.
\newblock Fully test-time adaptation for object detection.
\newblock In \emph{Proceedings of the IEEE/CVF Conference on Computer Vision and Pattern Recognition}, pages 1038--1047, 2024.

\bibitem[Sa et~al.(2016)Sa, Ge, Dayoub, Upcroft, Perez, and McCool]{sa2016deepfruits}
Inkyu Sa, Zongyuan Ge, Feras Dayoub, Ben Upcroft, Tristan Perez, and Chris McCool.
\newblock Deepfruits: A fruit detection system using deep neural networks.
\newblock \emph{sensors}, 16\penalty0 (8):\penalty0 1222, 2016.

\bibitem[Safonov et~al.(2025)Safonov, Bryntsev, Moskalenko, Kulikov, Vatolin, Timofte, et~al.]{ntire2025ugc}
Nickolay Safonov, Alexey Bryntsev, Andrey Moskalenko, Dmitry Kulikov, Dmitriy Vatolin, Radu Timofte, et~al.
\newblock Ntire 2025 challenge on {UGC} video enhancement: Methods and results.
\newblock In \emph{Proceedings of the IEEE/CVF Conference on Computer Vision and Pattern Recognition (CVPR) Workshops}, 2025.

\bibitem[Saleh et~al.(2020)Saleh, Laradji, Konovalov, Bradley, Vazquez, and Sheaves]{saleh2020realistic}
Alzayat Saleh, Issam~H Laradji, Dmitry~A Konovalov, Michael Bradley, David Vazquez, and Marcus Sheaves.
\newblock A realistic fish-habitat dataset to evaluate algorithms for underwater visual analysis.
\newblock \emph{Scientific Reports}, 2020.

\bibitem[Shangguan and Rostami(2023)]{shangguan2023identification}
Zeyu Shangguan and Mohammad Rostami.
\newblock Identification of novel classes for improving few-shot object detection.
\newblock In \emph{Proceedings of the IEEE/CVF International Conference on Computer Vision}, pages 3356--3366, 2023.

\bibitem[Shangguan and Rostami(2024)]{shangguan2024improved}
Zeyu Shangguan and Mohammad Rostami.
\newblock Improved region proposal network for enhanced few-shot object detection.
\newblock \emph{Neural Networks}, 180:\penalty0 106699, 2024.

\bibitem[Shao et~al.(2019)Shao, Li, Zhang, Peng, Yu, Zhang, Li, and Sun]{shao2019objects365}
Shuai Shao, Zeming Li, Tianyuan Zhang, Chao Peng, Gang Yu, Xiangyu Zhang, Jing Li, and Jian Sun.
\newblock Objects365: A large-scale, high-quality dataset for object detection.
\newblock In \emph{Proceedings of the IEEE/CVF international conference on computer vision}, pages 8430--8439, 2019.

\bibitem[Snell et~al.(2017)Snell, Swersky, and Zemel]{snell2017prototypical}
Jake Snell, Kevin Swersky, and Richard Zemel.
\newblock Prototypical networks for few-shot learning.
\newblock \emph{Advances in neural information processing systems}, 30, 2017.

\bibitem[Song and Yan(2013)]{song2013noise}
Kechen Song and Yunhui Yan.
\newblock A noise robust method based on completed local binary patterns for hot-rolled steel strip surface defects.
\newblock \emph{Applied Surface Science}, 2013.

\bibitem[Sun et~al.(2021)Sun, Li, Cai, Yuan, and Zhang]{sun2021fsce}
Bo Sun, Banghuai Li, Shengcai Cai, Ye Yuan, and Chi Zhang.
\newblock Fsce: Few-shot object detection via contrastive proposal encoding.
\newblock In \emph{CVPR}, 2021.

\bibitem[Sun et~al.(2025{\natexlab{a}})Sun, Alfarano, Duan, Su, Wang, Shi, Timofte, Paudel, Van~Gool, et~al.]{ntire2025event}
Lei Sun, Andrea Alfarano, Peiqi Duan, Shaolin Su, Kaiwei Wang, Boxin Shi, Radu Timofte, Danda~Pani Paudel, Luc Van~Gool, et~al.
\newblock Ntire 2025 challenge on event-based image deblurring: Methods and results.
\newblock In \emph{Proceedings of the IEEE/CVF Conference on Computer Vision and Pattern Recognition (CVPR) Workshops}, 2025{\natexlab{a}}.

\bibitem[Sun et~al.(2025{\natexlab{b}})Sun, Guo, Ren, Van~Gool, Timofte, Li, et~al.]{ntire2025denoising}
Lei Sun, Hang Guo, Bin Ren, Luc Van~Gool, Radu Timofte, Yawei Li, et~al.
\newblock The tenth ntire 2025 image denoising challenge report.
\newblock In \emph{Proceedings of the IEEE/CVF Conference on Computer Vision and Pattern Recognition (CVPR) Workshops}, 2025{\natexlab{b}}.

\bibitem[Tang et~al.(2022)Tang, Yuan, Li, and Tang]{tang2022learning}
Hao Tang, Chengcheng Yuan, Zechao Li, and Jinhui Tang.
\newblock Learning attention-guided pyramidal features for few-shot fine-grained recognition.
\newblock \emph{Pattern Recognition}, 130:\penalty0 108792, 2022.

\bibitem[Tseng et~al.(2020)Tseng, Lee, Huang, and Yang]{tseng2020cross}
Hung-Yu Tseng, Hsin-Ying Lee, Jia-Bin Huang, and Ming-Hsuan Yang.
\newblock Cross-domain few-shot classification via learned feature-wise transformation.
\newblock \emph{arXiv preprint arXiv:2001.08735}, 2020.

\bibitem[Vasluianu et~al.(2025{\natexlab{a}})Vasluianu, Seizinger, Zhou, Chen, Wu, Timofte, et~al.]{ntire2025shadow}
Florin-Alexandru Vasluianu, Tim Seizinger, Zhuyun Zhou, Cailian Chen, Zongwei Wu, Radu Timofte, et~al.
\newblock Ntire 2025 image shadow removal challenge report.
\newblock In \emph{Proceedings of the IEEE/CVF Conference on Computer Vision and Pattern Recognition (CVPR) Workshops}, 2025{\natexlab{a}}.

\bibitem[Vasluianu et~al.(2025{\natexlab{b}})Vasluianu, Seizinger, Zhou, Wu, Timofte, et~al.]{ntire2025ambient}
Florin-Alexandru Vasluianu, Tim Seizinger, Zhuyun Zhou, Zongwei Wu, Radu Timofte, et~al.
\newblock Ntire 2025 ambient lighting normalization challenge.
\newblock In \emph{Proceedings of the IEEE/CVF Conference on Computer Vision and Pattern Recognition (CVPR) Workshops}, 2025{\natexlab{b}}.

\bibitem[Wang et~al.(2020)Wang, Huang, Darrell, Gonzalez, and Yu]{wang2020frustratingly}
Xin Wang, Thomas~E Huang, Trevor Darrell, Joseph~E Gonzalez, and Fisher Yu.
\newblock Frustratingly simple few-shot object detection.
\newblock \emph{arXiv preprint arXiv:2003.06957}, 2020.

\bibitem[Wang et~al.(2023)Wang, Li, and Wu]{wang2023cardd}
Xinkuang Wang, Wenjing Li, and Zhongcheng Wu.
\newblock Cardd: A new dataset for vision-based car damage detection.
\newblock \emph{IEEE Transactions on Intelligent Transportation Systems}, 24\penalty0 (7):\penalty0 7202--7214, 2023.

\bibitem[Wang et~al.(2025)Wang, Liang, Zhang, Tian, Wang, Li, Yang, Timofte, Guo, et~al.]{ntire2025lightfield}
Yingqian Wang, Zhengyu Liang, Fengyuan Zhang, Lvli Tian, Longguang Wang, Juncheng Li, Jungang Yang, Radu Timofte, Yulan Guo, et~al.
\newblock Ntire 2025 challenge on light field image super-resolution: Methods and results.
\newblock In \emph{Proceedings of the IEEE/CVF Conference on Computer Vision and Pattern Recognition (CVPR) Workshops}, 2025.

\bibitem[Wolf et~al.(2019)Wolf, Debut, Sanh, Chaumond, Delangue, Moi, Cistac, Rault, Louf, Funtowicz, et~al.]{wolf2019huggingface}
Thomas Wolf, Lysandre Debut, Victor Sanh, Julien Chaumond, Clement Delangue, Anthony Moi, Pierric Cistac, Tim Rault, R{\'e}mi Louf, Morgan Funtowicz, et~al.
\newblock Huggingface's transformers: State-of-the-art natural language processing.
\newblock \emph{arXiv preprint arXiv:1910.03771}, 2019.

\bibitem[Wu et~al.(2020)Wu, Xia, and Wang]{wu2020adversarial}
Dongxian Wu, Shu-Tao Xia, and Yisen Wang.
\newblock Adversarial weight perturbation helps robust generalization.
\newblock \emph{Advances in neural information processing systems}, 33:\penalty0 2958--2969, 2020.

\bibitem[Xue et~al.(2024)Xue, Zheng, Fu, Ni, Zheng, Zhou, and You]{xue2024openmoe}
Fuzhao Xue, Zian Zheng, Yao Fu, Jinjie Ni, Zangwei Zheng, Wangchunshu Zhou, and Yang You.
\newblock Openmoe: An early effort on open mixture-of-experts language models.
\newblock \emph{arXiv preprint arXiv:2402.01739}, 2024.

\bibitem[Yang et~al.(2025)Yang, Cai, Ouyang, Vasluianu, Timofte, Ding, Sun, Fu, Li, Ho, Meng, et~al.]{ntire2025reflection}
Kangning Yang, Jie Cai, Ling Ouyang, Florin-Alexandru Vasluianu, Radu Timofte, Jiaming Ding, Huiming Sun, Lan Fu, Jinlong Li, Chiu~Man Ho, Zibo Meng, et~al.
\newblock Ntire 2025 challenge on single image reflection removal in the wild: Datasets, methods and results.
\newblock In \emph{Proceedings of the IEEE/CVF Conference on Computer Vision and Pattern Recognition (CVPR) Workshops}, 2025.

\bibitem[Zama~Ramirez et~al.(2025)Zama~Ramirez, Tosi, Di~Stefano, Timofte, Costanzino, Poggi, Salti, Mattoccia, et~al.]{ntire2025hrdepth}
Pierluigi Zama~Ramirez, Fabio Tosi, Luigi Di~Stefano, Radu Timofte, Alex Costanzino, Matteo Poggi, Samuele Salti, Stefano Mattoccia, et~al.
\newblock Ntire 2025 challenge on hr depth from images of specular and transparent surfaces.
\newblock In \emph{Proceedings of the IEEE/CVF Conference on Computer Vision and Pattern Recognition (CVPR) Workshops}, 2025.

\bibitem[Zha et~al.(2023)Zha, Tang, Sun, and Tang]{zha2023boosting}
Zican Zha, Hao Tang, Yunlian Sun, and Jinhui Tang.
\newblock Boosting few-shot fine-grained recognition with background suppression and foreground alignment.
\newblock \emph{IEEE Transactions on Circuits and Systems for Video Technology}, 33\penalty0 (8):\penalty0 3947--3961, 2023.

\bibitem[Zhang et~al.(2022)Zhang, Song, Gao, and Shen]{zhang2022free}
Ji Zhang, Jingkuan Song, Lianli Gao, and Hengtao Shen.
\newblock Free-lunch for cross-domain few-shot learning: Style-aware episodic training with robust contrastive learning.
\newblock In \emph{Proceedings of the 30th ACM international conference on multimedia}, pages 2586--2594, 2022.

\bibitem[Zhang et~al.(2023)Zhang, Liu, Wang, and Boularias]{zhang2023detect}
Xinyu Zhang, Yuhan Liu, Yuting Wang, and Abdeslam Boularias.
\newblock Detect everything with few examples.
\newblock \emph{arXiv preprint arXiv:2309.12969}, 2023.

\bibitem[Zhuo et~al.(2022)Zhuo, Fu, Chen, Cao, and Jiang]{zhuo2022tgdm}
Linhai Zhuo, Yuqian Fu, Jingjing Chen, Yixin Cao, and Yu-Gang Jiang.
\newblock Tgdm: Target guided dynamic mixup for cross-domain few-shot learning.
\newblock In \emph{Proceedings of the 30th ACM International Conference on Multimedia}, pages 6368--6376, 2022.

\bibitem[Zhuo et~al.(2024)Zhuo, Fu, Chen, Cao, and Jiang]{zhuo2024unified}
Linhai Zhuo, Yuqian Fu, Jingjing Chen, Yixin Cao, and Yu-Gang Jiang.
\newblock Unified view empirical study for large pretrained model on cross-domain few-shot learning.
\newblock \emph{ACM Transactions on Multimedia Computing, Communications and Applications}, 20\penalty0 (9):\penalty0 1--18, 2024.

\end{thebibliography}
